\pdfoutput=1

\documentclass[11pt]{article}

\usepackage[]{ACL2023}

\usepackage{times}
\usepackage{latexsym}

\usepackage[T1]{fontenc}

\usepackage[utf8]{inputenc}

\usepackage{microtype}

\usepackage{inconsolata}

\newcommand{\ei}{\textsc{Entity Inferences}}
\newcommand{\ecbd}{\textsc{ECBD}}

\usepackage[textsize=scriptsize]{todonotes}

\usepackage{enumitem}
\usepackage{booktabs}
\usepackage{multirow}
\usepackage{mathtools}
\usepackage{amsfonts}
\usepackage{amssymb}
\usepackage{fontawesome}
\usepackage[ruled,vlined]{algorithm2e}
\usepackage{arydshln}
\usepackage[procnames]{listings}
\usepackage{multicol}
\usepackage[title]{appendix}
\usepackage{graphicx} 
\usepackage{subcaption}
\usepackage{float}
\usepackage{colortbl}
\usepackage{fdsymbol}
\usepackage[T1]{fontenc} 
\newcommand\resulttablefontsize{\fontsize{7.7pt}{9.24pt}\selectfont}
\newcommand\exampletablefontsize{\fontsize{8.6pt}{10.32pt}\selectfont}
\title{Can LMs Learn New Entities from Descriptions? \\ Challenges in Propagating Injected Knowledge}

\author{
Yasumasa Onoe,
Michael J.Q. Zhang,
Shankar Padmanabhan,
Greg Durrett,
Eunsol Choi\\
Department of Computer Science\\
The University of Texas at Austin \\
{\tt\ yasumasa@utexas.edu}}

\begin{document}
\maketitle
\begin{abstract}
Pre-trained language models (LMs) are used for knowledge intensive tasks like question answering, but their knowledge gets continuously outdated as the world changes. Prior work has studied targeted updates to LMs, injecting individual facts and evaluating whether the model learns these facts while not changing predictions on other contexts. We take a step forward and study LMs' abilities to make inferences based on injected facts (or \emph{propagate} those facts): for example, after learning that something is a TV show, does an LM predict that you can watch it? We study this with two cloze-style tasks: an existing dataset of real-world sentences about novel entities (ECBD) as well as a new controlled benchmark with manually designed templates requiring varying levels of inference about injected knowledge. Surprisingly, we find that existing methods for updating knowledge (gradient-based fine-tuning and modifications of this approach) show little propagation of injected knowledge. These methods improve performance on cloze instances only when there is lexical overlap between injected facts and target inferences. Yet, prepending entity definitions in an LM's context improves performance across all settings, suggesting that there is substantial headroom for parameter-updating approaches for knowledge injection.

\end{abstract}

\section{Introduction}
Pre-trained language models (LMs) acquire comprehensive real-world knowledge from massive amounts of pre-training data, allowing them to use this knowledge effectively in downstream tasks. However, without continual updating, the knowledge contained within these backend LMs will eventually become outdated. This temporal mismatch affects model performance on downstream tasks \citep{zhang-choi-2021-situatedqa, Bhuwan_Dhingra_21, Angeliki_Lazaridou_21,jang-etal-2022-continual}. As LMs become more widely deployed, their knowledge should be synced with the current state of the world while maintaining reasonable deployment costs.

\begin{figure}[t]
    \centering
    \includegraphics[width=\linewidth]{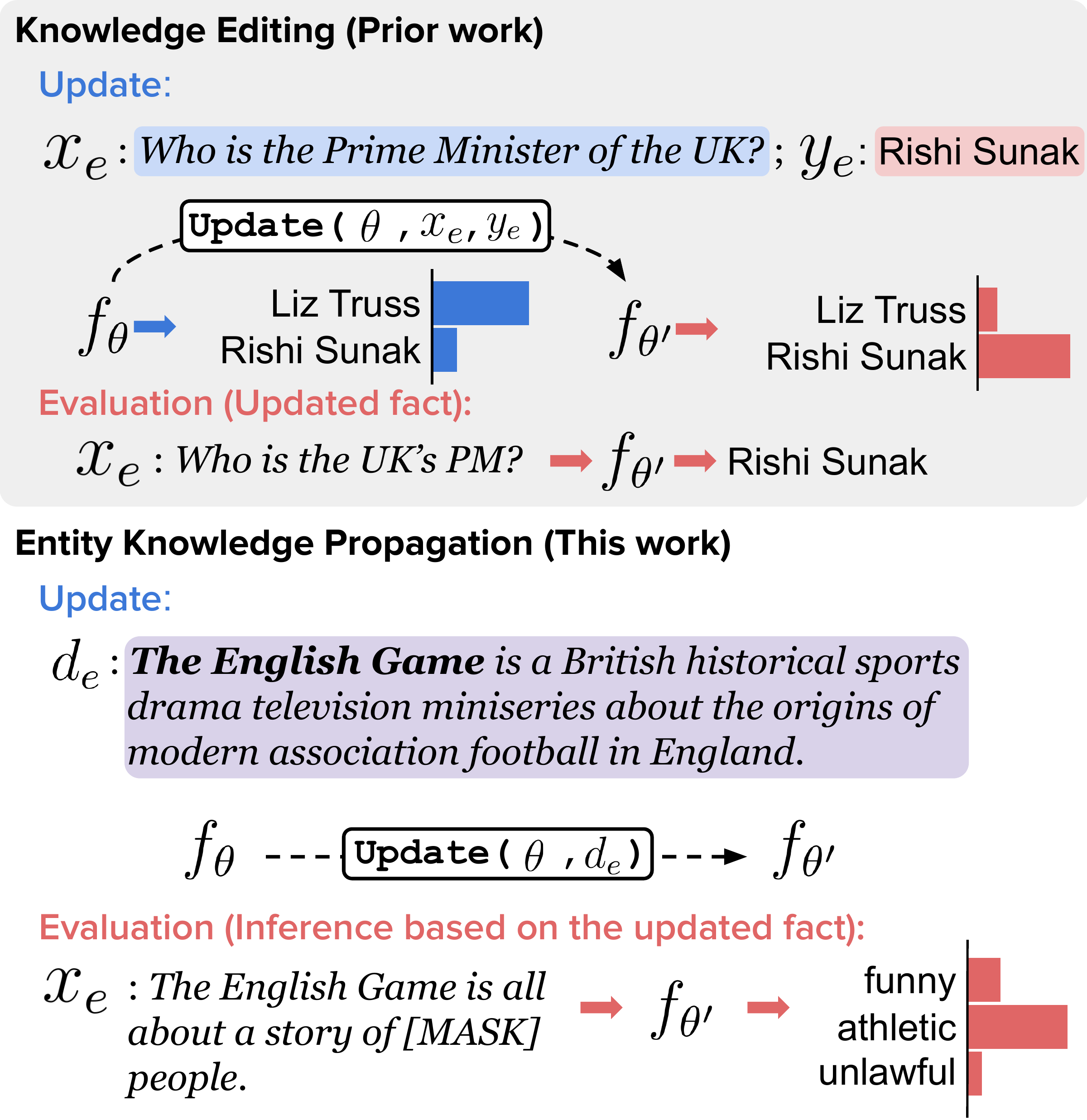}
    \vspace{-15pt}
    \caption{Knowledge editing tasks. We study a challenging \textbf{entity knowledge propagation} task where language models should make inferences after learning entities from their definitions. This differs from past knowledge editing which evaluates paraphrases of injected facts.}
    \label{fig:intro}
    \vspace{-15pt}
\end{figure}

Prior work has investigated knowledge editing in pre-trained LMs, updating model parameters to alter outputs to match what users want \citep{Chen_Zhu_2020, Anton_Sinitsin_20,Nicola_De_Cao_21_KE, mend, rome}.
In these studies, the original fact and the altered fact are provided (e.g., changing ``X was born in Y.'' to ``X was born in Z.''), and models are evaluated after a single update on each instance; see Figure~\ref{fig:intro} for an example. These model editing methods successfully provide targeted updates, fixing incorrect or outdated individual facts. Yet, \textbf{can LMs make inferences based on updated knowledge?} Prior evaluation has largely focused on two aspects of knowledge editing, whether the edits were successfully injected and whether other irrelevant sentences were impacted, but do not capture whether the LMs now can reason based on the new fact that has been injected. 

We take a step further and evaluate whether LMs can \emph{propagate} updated knowledge about new entities. We first inject definitions about the entity into LMs using various knowledge editing methods~\cite{mend,rome}, then evaluate LMs' performance on cloze tasks on a wide range of sentences about the entity (see Figure~\ref{fig:intro} for an example). We refer to this task as \emph{entity knowledge propagation} and introduce two cloze datasets to evaluate this challenging task.



Our first evaluation benchmark is the Entity Cloze By Date (\textsc{ECBD}) dataset \cite{onoe-etal-2022-entity}, which presents novel entities tagged with origination dates (e.g., \emph{Hurricane Ian}, 2022), their definition and probe sentences taken from their Wikipedia page. The task is to fill a masked span in probe sentences. Because Wikipedia contains a wide range of information, much of it not inferable from an entity's definition, injecting entity knowledge via its definition has an unclear impact on the probe sentences; filling in the masked span is nontrivial even after the entity definition is provided. For more controlled study, we introduce a new benchmark (\ei) with manually designed probe sentences with multiple-choice answer options. Once one learns about the definition of an emerging entity, finding the correct answer for these probe sentences is easy. 


We find that existing parameter updating methods can handle simpler inferences in \ei{}, but fail to improve performances in \ecbd, revealing a limitation in these methods. We further analyze the impact of fine-tuning. Distressingly, we find that 
simply prepending information in-context works very well, and matching the performance of this via parameter updates is challenging. A deeper analysis finds that model editing shows promising results only when the injected definition sentence and the cloze inference have lexical overlap. Our work establishes an evaluation paradigm and opens doors for work on editing methods that can propagate entity knowledge. 

\renewcommand{\arraystretch}{1}
\begin{table*}[t]
\centering
\exampletablefontsize
\begin{tabular}{lllll}
\toprule
Dataset & Entity ($e$) & Definition ($d_e$) & Probe Sentence ($x_e$) & Gold Span ($y_e$ / \{$C_y$\}) \\ \midrule
\multirow{2}{*}{\shortstack[l]{\sc Entity  \\ \sc Inferences}} & \multirow{2}{*}{\sf Dracula} &  Dracula is a drama horror television  & Dracula makes me & \texttt{scared} / \{ {\it athletic,} \\
& & serial developed by Mark Gatiss... &  feel <MASK>. & {\it brave, emotional, ...} \} \\ \midrule
\multirow{3}{*}{\sc ECBD} & \multirow{3}{*}{\sf Brexit} & Brexit was the withdrawal of the  & Studies estimate that Brexit \\ 
& & United Kingdom (UK) from the & and the end of <MASK> & \texttt{free movement} \\
& &  European Union (EU) at 23:00... & will likely result in a large...\\ \midrule
\multirow{3}{*}{\sc ECBD-Easy} & \multirow{3}{*}{\shortstack[l]{\sf Magnum  \\ \sf Fire}} & The Mangum Fire was a wildfire. & On June 14, the Mangum Fire \\
& & burning in Kaibab National Forest & jumped control lines towards & \texttt{Arizona} \\
& & in Arizona in the United States. &  Mangum Springs, <MASK>... \\
\bottomrule
\end{tabular}

\caption{Examples from each dataset outlined in Section~\ref{sec:datasets}. Unlike ECBD and ECBD-Easy, the gold spans in Entity Inferences examples are always one of several multiple-choice options per example.}  \label{tab:examples}
\vspace{-7pt}
\end{table*}

\section{Entity Knowledge Propagation}

We propose \emph{Entity Knowledge Propagation (EKP)}\footnote{ The code and data are available at \url{https://github.com/yasumasaonoe/entity_knowledge_propagation}.}, a new task where we want to update model parameters to reflect an emerging entity that is unseen in the LMs' pre-training corpus. For example, BERT was trained in 2018, so COVID-19 is an emerging entity to BERT. We explore various ways of editing model parameters based on definition sentences to inject new knowledge. Once we inject the knowledge of the emerging entity into the model parameters, we evaluate the updated model's ability to reason about the emerging entity. 

\subsection{Task Definition}
\label{sec:task_definition}

Formally, we have a language model $f_{\theta}$ with parameters $\theta$. An input to the model consists of a (partial) sentence or chunk of text $x_e$ that contains at least one explicit reference to an emerging entity $e$ (i.e., invoking $e$ by name). We use $f_{\theta}(y_e \mid x_e)$ to denote placing a probability distribution over a text sequence $y_e$ given the text $x_e$.\footnote{In autoregressive models, $y_e$ can be a continuation of $x_e$; in mask filling models like T5 or BART, $x_e$ can contain mask tokens and $y_e$ consists of those mask fillers.}

Our data instances have the property that $y_e$ represents an \emph{inference} we make about the entity: $y_e$ must be related to the entity $e$ such that an LM should give higher probability to it if the LM ``knows'' $e$ well. We do not expect the raw model $f_{\theta}$ to perform well without any updates, since the entity $e$ is completely unseen during the pre-training stage. We assume that the emerging entity comes with a short \emph{definition sentence} $d_e$ that provides basic information about the entity. This provides the basis for the update to $f_\theta$.

To summarize, each example $\langle e, d_e, x_e, y_e \rangle \in \mathcal{D}$ consists of an emerging entity $e$, a definition sentence $d_e$, a probe sentence $x_e$, and a gold completion $y_e$. Knowledge editing methods will compute $\theta' \leftarrow \mathrm{update}(\theta, e, d_e)$, updating parameters $\theta$ regarding $e$ and its definition $d_e$, to give higher probability for future inferences about $e$ like those expressed by $x_e$ and $y_e$ (examples in Figure~\ref{fig:intro}).
\paragraph{Metrics}
Following prior work in the knowledge updating literature~\cite{Chen_Zhu_2020, Nicola_De_Cao_21_KE, mend, rome}, we will evaluate two criteria: update success and specificity. Each of these criteria is evaluated with respect to a base metric, which is either perplexity or accuracy, depending on our dataset. We will define them here in the case of perplexity (lower is better); we will use the same definitions for accuracy, but the desired trends will be opposite.

For update success, we will measure if the perplexity of the updated model $\mathrm{ppl}(f_{\theta'}(y_e \mid x_e))$ is better than the raw model $\mathrm{ppl}(f_{\theta}(y_e \mid x_e))$ (lower perplexity is better). For specificity, we compute the difference between the post-update perplexity and pre-update perplexity $\mathrm{ppl}(f_{\theta'}(y_{\hat{e}} \mid x_{\hat{e}})) - \mathrm{ppl}(f(y_{\hat{e}} \mid x_{\hat{e}}))$ for $\hat{e} \neq e$, entities other than $e$. Ideally, we want this perplexity value to be close to zero; a positive value indicates that perplexity has gotten worse on these entities after the update. It can theoretically be negative if the update makes the LM to guess irrelevant examples better.

\paragraph{Comparison with prior tasks} Similar editing procedures have been explored in the literature, but with key differences from our setting. A line of work on \textbf{knowledge editing} \cite{Chen_Zhu_2020, Nicola_De_Cao_21_KE, mend} addresses a version of our task where $f_\theta$ is updated to encode information about a particular fact. This could be written as $\theta' \leftarrow \mathrm{update}(\theta, x_e, y)$. They then evaluate $f_\theta(y \mid \tilde{x}_e)$ on perturbed inputs $\tilde{x}_e$ that are paraphrases of the $x_e$ they inject. The answer $y$ is visible when the network is updated and it simply needs to be preserved for future (paraphrased) queries. By contrast, in our setting, $y$ and the injected definition $d_e$ may have little overlap. 

ROME \cite{rome} addresses knowledge editing as well as a variant of \textbf{counterfactual model editing}. This task involves an update similar in spirit to ours: $\theta' \leftarrow \mathrm{update}(\theta, e, (x_{e,1}, y_{e,1}))$ that updates a completion of a sentence (e.g., $x_{e,1}=$\emph{the Eiffel Tower is located in}, $y_{e,1}=$\emph{Rome}) and then expects the knowledge to be usable for other inference pairs $(x_{e,2}, y_{e,2})$. These differ in that the injected knowledge is not a complete definition of an entity; while their method could theoretically be used for our task, it relies on localizing and editing \emph{existing} information about $e$. Therefore, it is less appropriate in handling emerging entities, as our results will show. 



\section{Constructing benchmarks for EKP}
\label{sec:datasets}

\begin{table}
\small
\begin{center}
\begin{tabular}{lrrr}
\toprule
Dataset & \# Examples & \# Entities  & $y_e$ in $d_e$  \\\midrule
Entity Inferences & 170 & 85 &  92  \\
ECBD & 1000  & 208 & 29   \\
ECBD-easy & 152 & 74 & 152 \\
\bottomrule
\end{tabular}
\end{center}
\caption{Statistics from each EKP dataset. We report the number of examples in each evaluation set in addition to the number of unique entities and total instances where the gold span can be found within the entity description.}\label{tab:dataset_statistics}
\end{table}

We use two different types of datasets to investigate how new entity knowledge is propagated into the LM's parameter space. Table~\ref{tab:dataset_statistics} summarizes the dataset statistics on two benchmarks, including the extent to which the target spans $y$ overlap with the definitions $d_e$, which will be important later.

\subsection{\textbf{\textsc{ECBD}}}  
Entity Cloze By Date \cite[ECBD]{onoe-etal-2022-entity} presents entities indexed by their origination dates paired with probe sentences containing those entities. In addition, the dataset provides the definition sentence (first sentence sentence of Wikipedia article) for each entity. The original task focuses on general temporal adaptation of language models, evaluating model's perplexity in predicting masked spans in probe sentences. We repurpose this dataset to focus on \textbf{targeted} knowledge updates and the propagation of entity knowledge. We take entities with origination date between 2020/01 and 2021/09 to ensure they are unseen by the LMs we study.


These instances fall into the paradigm discussed in Section~\ref{sec:task_definition} (example shown in Table~\ref{tab:examples}):

\begin{description}[noitemsep,topsep=0pt,leftmargin=*]
\item [$e: \tt{Entity}$]: the title of the Wikipedia article
\item [$d_e: \tt{Definition Sentence}$]: the first sentence of the Wikipedia article for the entity.
\item [$x_e: \tt{Probe Sentence}$]: a sentence selected from the Wikipedia article according to the procedure described in \citet{onoe-etal-2022-entity}
\item [$y: \tt{Gold Span}$]: the target span as described in \citet{onoe-etal-2022-entity}
\end{description}

\paragraph{\textsc{ECBD-easy}} We filter ECBD to create ECBD-easy, a subset where knowledge propagation should be easier. Specifically, we take cases where the target masked span $y$ is contained in the definition sentence $d_e$ verbatim; such examples are more congruent with the formulation of past work such as MEND and are typically easier, as simply boosting the probability of the definition tokens can improve perplexity on the gold span.

\paragraph{Evaluation Metrics} Following \citet{onoe-etal-2022-entity}, we compute per-token perplexity over the masked spans. Because of differences in model architecture such as tokenizer choice, this metric does not allow comparison across different base models. We randomly sample 40 entities as $\hat{e}$ from ECBD popular subset to measure specificity.


\subsection{\textbf{\textsc{Entity Inferences}}}
While ECBD contains real-world sentences spanning a broad domain, it presents a very challenging task even for humans, often requiring rich knowledge and various types of reasoning and inference. For a more controlled study targeting on knowledge propagation, we construct a new dataset we name as \textsc{Entity Inferences}. We describe the data construction process in Appendix~\ref{app:entit_inference_data_constuction}. 

In this dataset, choosing the correct span is much easier when given the definition sentence.
Further, instead of requiring LMs to predict spans from open vocabulary, we provide a set of candidate spans and evaluate whether LMs can assign higher probability to the correct answer candidate. Instances here are designed to be similar to ECBD, but the probe sentences $x_e$ are handcrafted to elicit the target inference type, and the gold span $y$ comes with an associated set $\{C_y\}$ of options. There can be two inference types: explicit and implicit. The explicit probe sentences ask information that is explicitly stated in the definition sentence (e.g., the genre of a TV show). On the other hand, the implicit probe sentences require commonsense-like information (e.g., people \emph{watch a TV show}, and they don't \emph{eat a TV show}.).

\paragraph{Evaluation metrics} For this multiple-choice cloze task, we evaluate knowledge propagation by meausring \textbf{accuracy} (i.e., how often the gold label gets the highest probability over all answer candidates).
In addition, we compute the \textbf{specificity score} by evaluating a model on other probe sentences from similar entities.
 
\section{Experimental Setup}\label{sec:experiments}



\subsection{Base Language Models} 
Model architectures can have impact on their capabilities of acquiring entity knowledge. Thus, we consider both left-to-right and seq-to-seq model architectures. Specifically, we use GPT-Neo 1.3B~\cite{gpt-neo}\footnote{GPT-Neo has been trained on the Pile dataset \citep{pile}, which is collected in 2020.} and T5-large~\cite{T5}\footnote{T5 has been trained in 2019 on the C4 dataset.} as base language models ($f_{\theta}$), available via Huggingface Transformers \citep{wolf-etal-2020-transformers}. We additionally consider GPT2-XL \citep{radford2019language} as a base model to closely follow the protocol presented in ROME paper~\cite{rome}.

\subsection{Parameter Updating Methods}


\paragraph{Finetuning} is a common way for adapting a pre-trained LM to a specific task or domain \cite{gururangan-etal-2020-dont}.
In a similar vein, we aim to adopt a pretrained LM to an environment where new entities constantly arise.
Given $e$ and its definition $d_e$, we update the parameters $\theta$ to minimize loss on a training example formed from $d_e$.
For left-to-right models (e.g., GPT-Neo), we use the standard next token prediction language modeling task on the entire $d_e$ example.
For mask filling models (T5), we randomly select a span\footnote{We uniformly draw span length between 1 and 5 to match the average span length used during pretraining of T5.} that is not overlapping with the entity mention span, following \citet{onoe-etal-2022-entity}. We experiment with two fine-tuning settings: \textbf{full model} (updating all parameters) and \textbf{last layer} (updating parameters belonging to the last transformer layer only). We start finetuning from the original model checkpoint for each example.\footnote{\citet{Chen_Zhu_2020} reports that finetuning on an individual fact (FTM) constantly outperforms finetuning with mixed facts (FTA).}




\renewcommand{\arraystretch}{1}
\begin{table*}[t]
	\centering
	\resulttablefontsize
	\setlength{\tabcolsep}{4pt}
	\begin{tabular}{l l c c c c c c }

		\toprule
		\multicolumn{2}{c}{} & \multicolumn{2}{c}{\textsc{Entity Inferences} (Accuracy)}  & \multicolumn{2}{c}{\textsc{ECBD} (Perplexity)}  & \multicolumn{2}{c}{\textsc{ECBD-easy}  (Perplexity)}\\
	    \cmidrule(r){3-4}  \cmidrule(r){5-6}   \cmidrule(r){7-8} 
		\multicolumn{2}{c}{Method} &  \multicolumn{1}{c}{ Target ($\Delta$)} & \multicolumn{1}{c}{Specificity ($\Delta$)}  & \multicolumn{1}{c}{ Target ($\Delta$)} & \multicolumn{1}{c}{Specificity ($\Delta$)} & \multicolumn{1}{c}{Target ($\Delta$)} & \multicolumn{1}{c}{Specificity ($\Delta$)} \\
        \midrule
		\multicolumn{1}{l}{\scriptsize \textcolor{darkgray}{Type: left-to-right}} &\multicolumn{6}{c}{\textbf{\small GPT-Neo}}  & \multicolumn{1}{r}{\scriptsize \textcolor{darkgray}{Size: 1.3B}} \\
		\midrule
		\multirow{4}{*}{Model Editing} & Base Model & 34.1 & 34.1  & 38.8 & 26.1 & 21.1 & 26.1    \\
		& FT (full model) & 57.7 ($+$23.6)  & 18.3 ($-$15.9)  & 36.8 ($-$2.0) &  26.0 ($+$0.1) & 12.1 ($-$9.0) & 26.0 ($-$0.1)    \\
		& FT (last layer) &  48.8 ($+$14.7) &  16.4 ($-$17.7)  & 38.7 ($-$0.1) &  26.0 ($+$0.1) & 19.6 ($-$1.5) & 26.1 (0.0)\:\:\:  \\
		& MEND & 41.8 ($+$7.7)\:\:  & 34.4 ($+$0.3)\:\:  & 48.6 ($+$9.8) & 27.2 ($+$1.1) & 12.6 ($-$8.5) & 28.1 ($+$2.1)\\
		\midrule
        \multirow{2}{*}{Input Augmentation} & Definition &{ 60.0 ($+$25.9)}  &\textit{ 34.1} & 22.5 ($-$16.3)  & \textit{ 26.1}  & \:\:\:\:3.2 ($-$17.9)  & \textit{26.1}  \\
        & Random Def. & 27.7 ($-$6.4)\:\:  &\textit{ 34.1} & 55.1 ($+$16.3)  & \textit{ 26.1}  &  \:\:35.7 ($+$14.6) &  \textit{26.1} \\
		\midrule
		\multicolumn{1}{l}{\scriptsize \textcolor{darkgray}{Type: seq-to-seq}} & \multicolumn{6}{c}{\textbf{\small T5 Large}} & \multicolumn{1}{r}{\scriptsize \textcolor{darkgray}{Size: 770M}} \\
		\midrule
		\multirow{4}{*}{Model Editing} & Base Model & 42.9 &  42.9  & 17.0 &   12.9  & 14.3 & 12.9   \\
		& FT (full model) & 64.7 ($+$21.8)  & 38.2 ($-$4.7) & 17.0 (0.0)\:\:\:  &  12.9 (0.0)  & 14.3 (0.0)\:\:\: & 12.8 ($-$0.1)    \\
		& FT (last layer) & 52.9 ($+$10.5) & 43.9 ($+$1.0)  & 17.0 (0.0)\:\:\:  &  12.9 (0.0)  & 14.2 ($-$0.1) & {12.9} (0.0)\:\:\:   \\
		& MEND & 43.5 ($+$0.6)\:\:  & 42.7 ($-$0.2) & 17.3 ($+$0.3) & 12.9 (0.0) & 14.0 ($-$0.3) & 12.9 (0.0)\:\:\:  \\
		\midrule
        \multirow{2}{*}{Input Augmentation} & Definition & 73.5  ($+$30.6)  & \textit{42.9} & 12.4  ($-$4.6)  & \textit{12.9} & 13.6 ($-$0.7) & \textit{12.9}   \\
        & Random Def. & 42.4  ($-$0.5)\:\:  & \textit{42.9} & 15.8  ($-$1.2)  & \textit{12.9}  & 13.6 ($-$0.7) & \textit{12.9} \\
		\midrule
		\multicolumn{1}{l}{\scriptsize \textcolor{darkgray}{Type: left-to-right}} &\multicolumn{6}{c}{\textbf{\small GPT2-XL}}  & \multicolumn{1}{r}{\scriptsize \textcolor{darkgray}{Size: 1.5B}} \\
		\midrule
		\multirow{4}{*}{Model Editing} & Base Model & 32.9 & 32.9  & 42.8 & 25.4    & 31.0  &  25.4  \\
        & FT (full model) & 62.9 ($+$30.0) & 24.1 ($-$8.8) & 39.4 ($-$3.4) & 25.4 (0.0) & 16.8 ($-$14.2) & 25.4 (0.0)  \\
		& FT (last layer) & 46.5 ($+$13.6) & 35.4 ($+$2.5) & 42.8 (0.0)\:\:\: & 25.4 (0.0)  & 30.4 ($-$0.6)\:\: & 25.4 (0.0)  \\
	    & ROME & 54.3 ($+$23.5)  & 29.9 ($-$2.0) & N/A & N/A & N/A & N/A   \\
        \midrule
        \multirow{2}{*}{Input Augmentation} & Definition &  64.1 ($+$31.2) & \textit{32.9}  & 26.6 ($-$16.2)  &   \textit{25.4} & \:\:\:\:3.5 ($-$27.5) & \textit{25.4} \\
        & Random Def. & 26.5 ($-$6.4)\:\:   & \textit{32.9}   & 56.3 ($+$13.5)  & \textit{25.4}   & 37.1 ($+$6.1)  & \textit{25.4}  \\
		\bottomrule
	\end{tabular}
	\caption{Evaluation results. On \textsc{Entity Inferences}, both fine-tuning and ROME show large increases in accuracy with various costs to specificity, although MEND is ineffective. On the more challenging \textsc{ECBD} data, despite Input Augmentation suggesting that knowledge is relevant, no technique leads to a decrease in perplexity, although we do see some gains on \textsc{ECBD-easy}.} \label{tab:main-results}
\end{table*}

\paragraph{MEND~\cite{mend}} can be viewed as a hypernetwork that efficiently transforms the raw finetuning gradient into a parameter update that should successfully edit the base model's parameters in one step. This method is designed for injecting or editing individual facts about entities, not a collections of facts about entities (i.e., a complete definition's worth of entity knowledge). The MEND parameters are trained on an editing dataset where each example consists of an input-output pair, an altered output, and locality examples (for measuring sensitivity). The goal of MEND training is to learn a network that modifies the target fact without affecting unmodified facts. 

We train MEND editors for GPT-Neo and T5 with the WikiText-103 dataset, which uses generated text as altered output following the configuration used in the original paper.\footnote{\url{https://github.com/eric-mitchell/mend}}

\paragraph{ROME~\cite{rome}} performs knowledge editing by treating a MLP as a key-value storage: it uses a subject (such as the Eiffel Tower) to extract the ``value'' associated with that subject in the MLP. Then, it uses a rank-one modification of the weights of the MLP to ``rewrite'' this key-value pair. 

We use the ROME editor for GPT2-XL. We format according to the subject, relation, and object structure of ROME prompts; examples of these can be found in the Appendix. The subject is a one-word name 
of the entity, the relation is the definition sentence before the <MASK> token, and the object is the correct label. Examples in which the subject did not appear before the <MASK> token (less than 0.5\% of our data) were filtered.\footnote{Additionally, a number of examples in ECBD had a format incompatible with ROME; for example, ROME is currently unable to run on examples with special tokens such as '(' or '*' immediately surrounding the subject. We will discuss later the performance of ROME on ECBD and why we do not formally report it.}



\subsection{Input Augmentation}\label{subsec:prepend_def}

Finally, as was explored in \citet{onoe-etal-2022-entity}, we evaluate an approach where information is added only in-context: prepending a definition sentence to a probe sentence (\emph{Definition}). While such input augmentation will lower the efficiency (as the context length has increased) and will not yield an updated model, a lower perplexity can indicate if the definition sentence contains useful information and can show what gains are achievable. We also present a baseline that prepends a randomly chosen definition of another entity (\emph{Random Def.}), following prior work.


\subsection{Computational Cost}

While input augmentation is the simplest to implement out of all the knowledge injection methods we experiment with, it comes with an increased computational cost at inference time due to the longer input sequence. A principal goal of this line of work is to update models so they can learn about many new entities over time; therefore, we do not consider input augmentation a valid solution to the overall problem in this work due to poor scaling. 

In contrast, performing knowledge injection via finetuning carries the upfront cost of computing and performing gradient updates, but has no such cost increase during inference. Computing these gradient updates, however, can become quite burdensome when injecting many facts into the same LM. This, in part, is the motivation behind methods like MEND which have an additional upfront cost of training a meta-network to predict the necessary parameter updates. After training the meta-network, the amortized cost of updating many individual facts becomes much cheaper. While this dramatically reduces the cost of performing multiple edits to a single LM, meta-networks must be retrained for each unique LM we wish to update.

\section{Results}
Table~\ref{tab:main-results} reports the performances of various knowledge injection approaches on three base models. In all experimental setting, we see input augmentation (prepending definition) boasts robust and consistent performances gains. Prepending random definitions hurt performances in GPT-Neo while does not impact T5.
This indicates that the definition contains information relevant to the spans to predict.
As model behaves substantially differently across datasets, we first separately discuss the results on each dataset, \textsc{Entity Inferences}, \textsc{ECBD}, and \textsc{ECBD-easy}, and then draw larger conclusions.  
 
\subsection{\textsc{Entity Inferences}}
Here, we observe \textbf{fine-tuning is broadly effective at improving accuracy}. Finetuning (full model) brings up the post-edit accuracy by more than 20 points for all three base models. Yet, it comes at the cost of \textbf{medium to large decreases in specificity}, with drops of 15.9 and 8.8 points on GPT-Neo and GPT2-XL. MEND overall does not cause a substantial change in the model, as shown by the impact on specificity (+0.3). 
ROME does not achieve editing performance as strong as fine-tuning on GPT2-XL (+30 vs. +23.5), but it does so with a lower impact to specificity (-8.8 vs. -2.0).

On this benchmark, where evaluation metric is accuracy, we can make comparison across the models. Overall we see better performances with T5 model, despite it being the smallest model we test, potentially as it uses both left and right context. 



\subsection{\textsc{ECBD}}
On our most challenging benchmark setting, ECBD, \textbf{none of the model editing techniques, including fine-tuning, lead to substantial decrease in perplexity nor increase in specificity.} 
MEND even causes a increase in perplexity when the base model is GPT-Neo.

We attempted to evaluate ROME in this setting. However, we found very poor performance (perplexities of over 100 for both datasets). We do not report these in the table as technically ECBD is out of scope for ROME: ROME relies on a particular (entity, relation, object) format that is not well-suited to updating a model with general definitional knowledge of an entity, as opposed to specific attributes like \emph{The English Game is a drama} in \textsc{EntityInferences}. Attempting to force our definitions into ROME's expected form led to very high perplexities (over 100 on both ECBD sets).

These observation implies that the current model editing approaches are not able to propagate entity knowledge to the probe sentences just from the definition sentences. The inference patterns in the \textsc{ECBD} examples might be too complex to be effectively learned by a small number of parameter updates on a few examples, requiring implicit, multihop, and commonsense reasoning.  




\begin{figure*}
    \centering
    \begin{subfigure}{0.329\linewidth}
    \centering
    \includegraphics[width=\linewidth]{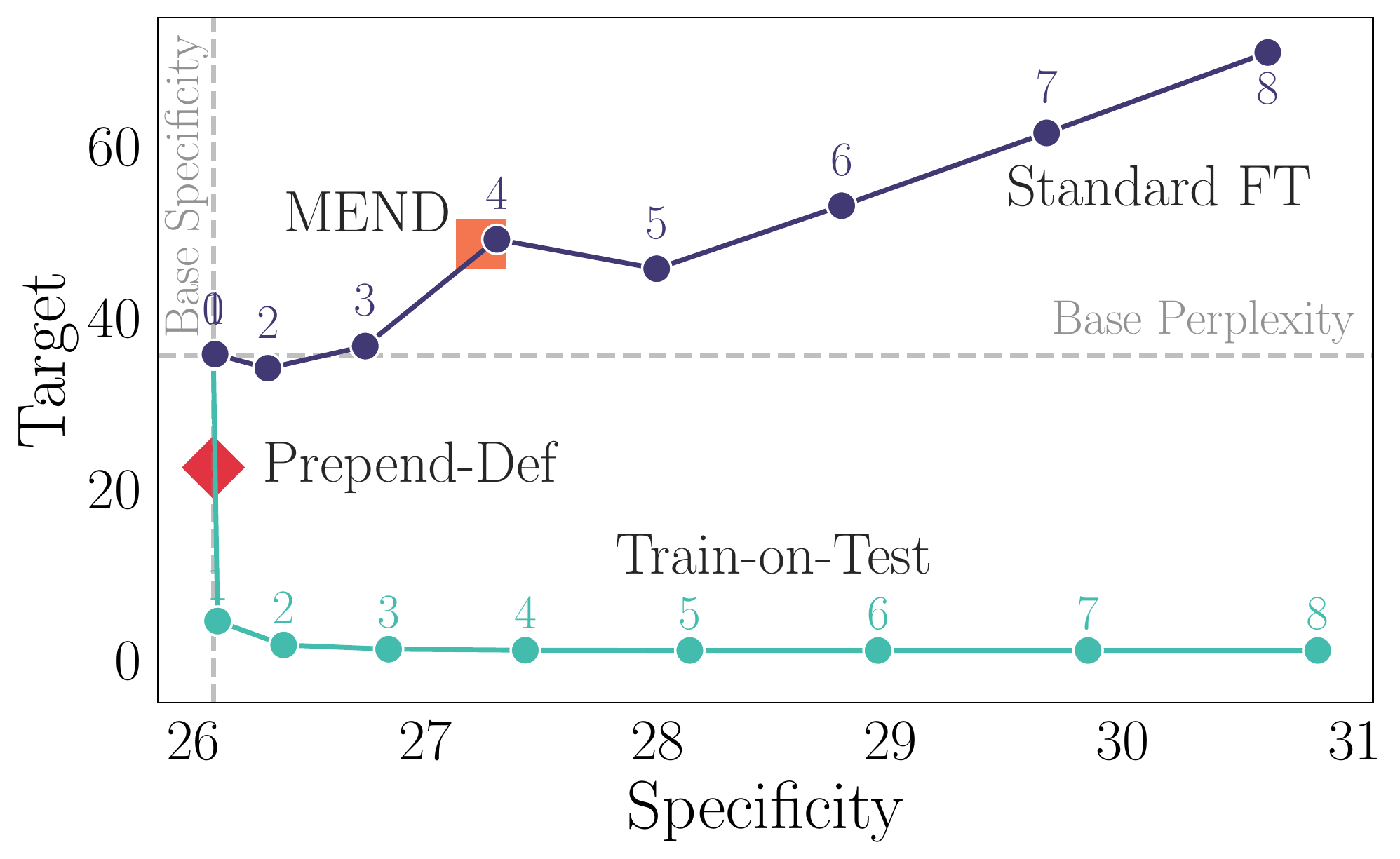}
    \caption{\textsc{ECBD}}
    \label{fig:tradeoff-ecbd}
    \end{subfigure}
    \begin{subfigure}{0.329\linewidth}
    \centering
    \includegraphics[width=\linewidth]{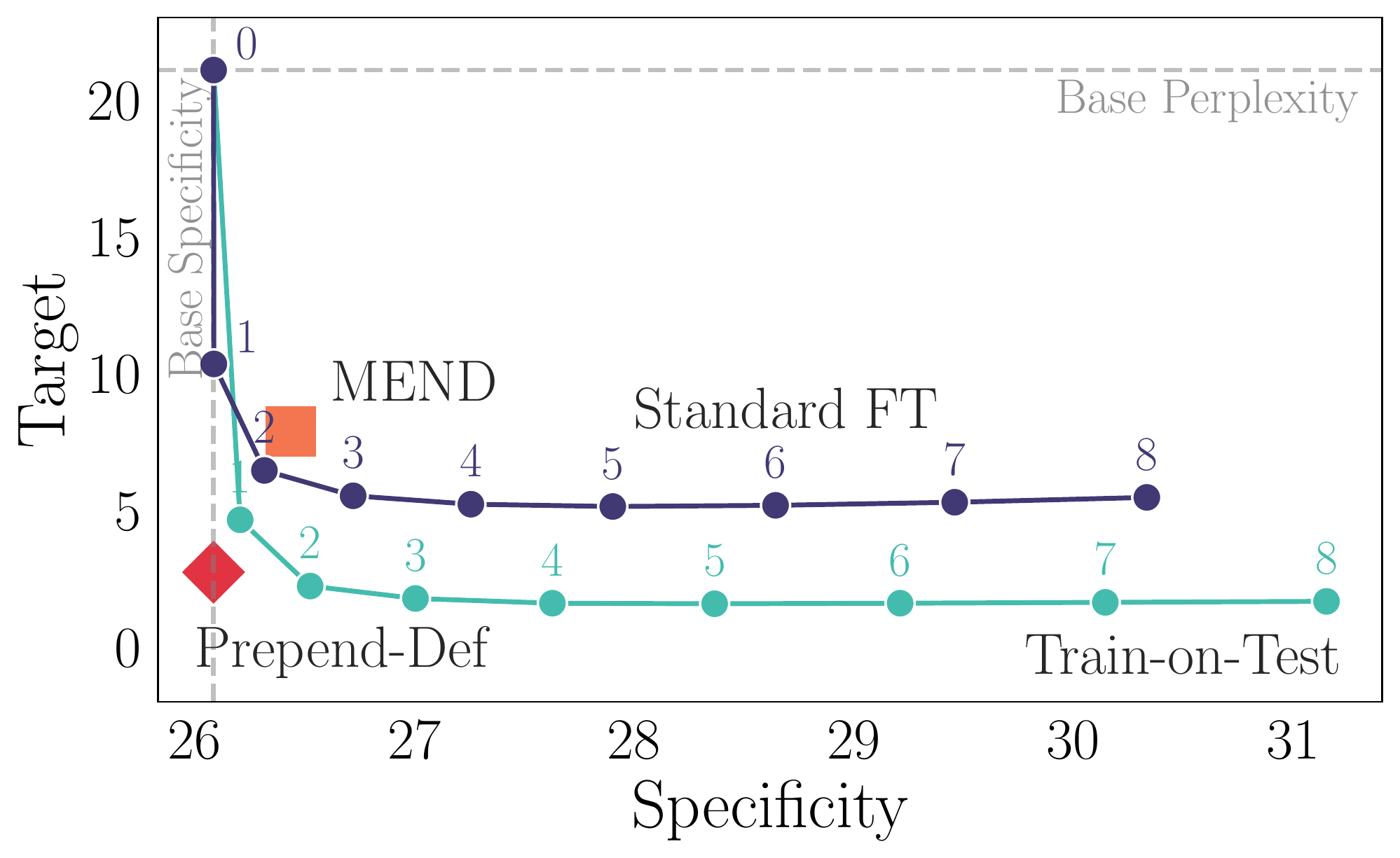}
    \caption{\textsc{ECBD-easy}}
    \label{fig:tradeoff-ecbd-easy}
    \end{subfigure}
    \begin{subfigure}{0.329\linewidth}
    \centering
    \includegraphics[width=\linewidth]{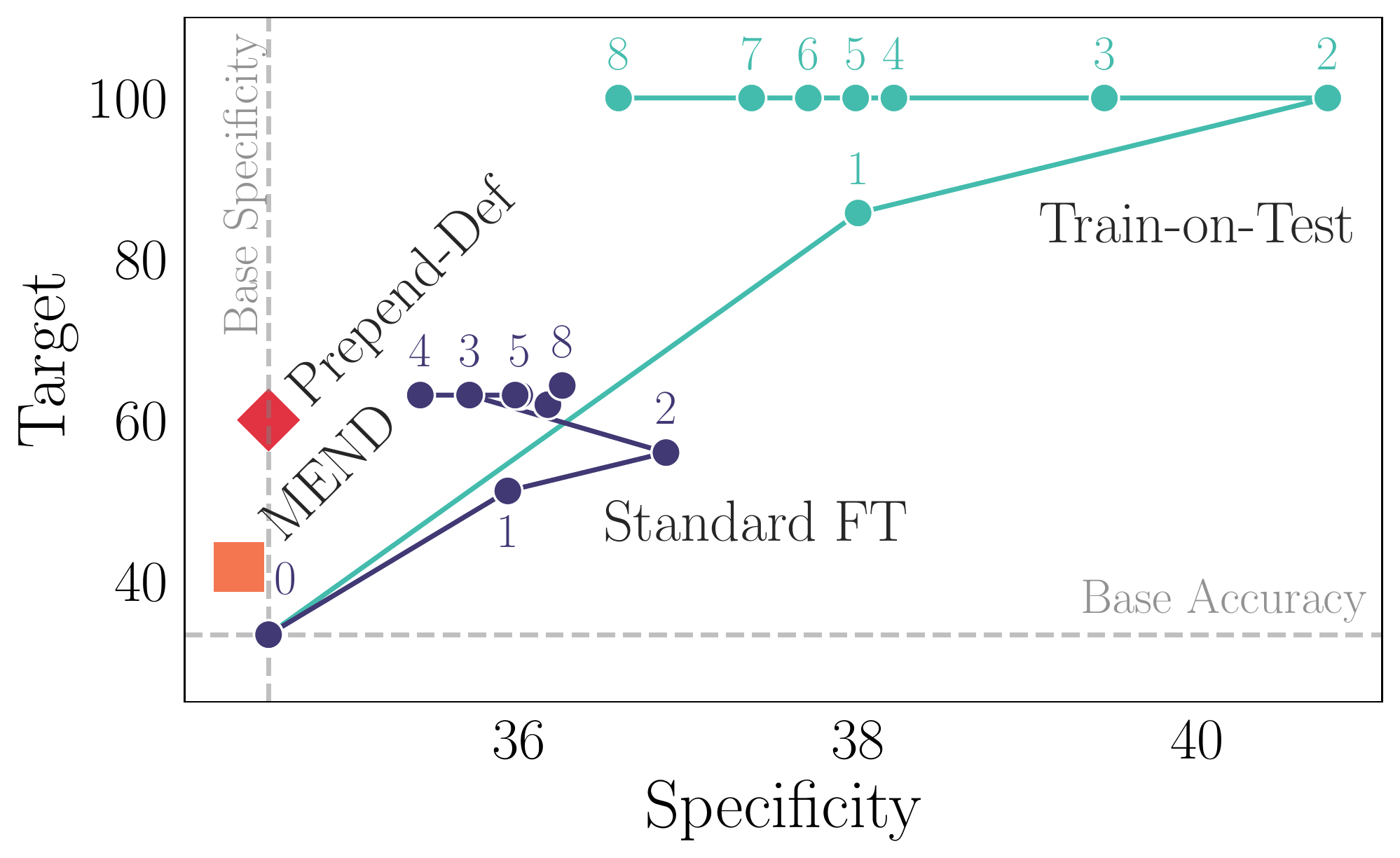}
    \caption{\textsc{Entity Inferences}}
    \label{fig:tradeoff-entity-inferences}
    \end{subfigure}
    \caption{The tradeoff curves of finetuning on (a) \textsc{ECBD} and (b) \textsc{ECBD-easy}. We plot perplexity ($y$-axis) and specificity ($x$-axis) measured with various numbers of epochs ranging from 0 to 8. For (c) \textsc{Entity Inferences}, higher is better because the $x$-axis is now accuracy. We plot MEND (\textcolor{orange} {$\blacksquare$}) and prepending definition (\textcolor{red}{$\medblackdiamond$}) as points. Fine-tuning and MEND can do nearly as well as train-on-test on \textsc{ECBD-easy}, but they fail dramatically on \textsc{ECBD}.}
    \label{fig:tradeoff}
\end{figure*}


\subsection{\textsc{ECBD-easy}}
To understand the low performance in the ECBD setting, we look more closely into \textsc{ECBD-easy} examples, where the gold spans are always included in the definition sentences. On this subset of \textsc{ECBD}, finetuning and MEND is effective on GPT-Neo, decreasing perplexity by $9.0$ and $8.5$ respectively. T5-large does not change its post perplexity. This is potentially because T5 only predicts and updates on masked spans (which might not contain the gold span), unlike the other two base models. 


Mildly positive results on the easier subset, along with robust performances of input augmentations, lead us to conclude that the gains are achievable. Yet, existing knowledge editing techniques may be restricted to reproducing the knowledge directly injected into the model. We launch a further investigation into what makes this task challenging. 

\section{Analysis}

We analyze the challenges in knowledge propagation by first estimating an informal upper bound of model editing performance (Section~\ref{subsec:upper_bound}).
We then examine how the similarity between the definition sentence and probe sentence impacts the performance of model editing (Section~\ref{subsec:info_overlap}), inspired by positive performances on \textsc{ECBD-easy} subset. We conduct our analysis with GPT-Neo base model on random subsets of \textsc{Entity Inferences} (half the data) and \textsc{ECBD} (100 NP span and 100 random span) to reduce computational costs.

\subsection{Targeted Update / Specificity Tradeoff}\label{subsec:upper_bound}

\paragraph{Performance Upper Bound} We estimate a performance upper bound for fine-tuning by setting {the definition and probe sentences to be identical}. In this case, sufficiently large gradient updates should lead to arbitrarily good performance from fine-tuning. We call this setting \emph{Train-on-Test}. 


For our three datasets (\textsc{Entity Inferences}, \textsc{ECBD}, and \textsc{ECBD-easy}), we finetune a model for a range of 1 to 8 epochs (i.e., the number of updates).
We use a learning rate of 5e-5 for \textsc{Entity Inferences} and plot the specificity score vs accuracy.
For \textsc{ECBD} and \textsc{ECBD-easy}, we choose a learning rate of 3e-5 and then compare the specificity score and perplexity. These learning rates were chosen to optimize performance from the range of values described in Appendix~\ref{app:hyperparams}.


\paragraph{Findings}
Figure~\ref{fig:tradeoff-ecbd} depicts the perplexity--specificity tradeoff curves of fine-tuning approach on \textsc{ECBD} dataset. The perplexity and the specificity score by the base model are drawn as the horizontal dotted line and the vertical dotted line respectively.
Ideally, we want a model to achieve low perplexity and the specificity score identical to the base score (performance in the lower left corner).
On ECBD, we see that \emph{Standard FT} shows an upward trend: with larger parameter updates, we worsen the specificity as expected, but also perplexity, meaning that finetuning for longer does not usefully propagate entity information from the definition sentence into the model.
Input augmentation (Prepend-Def) performs robustly, indicating that the issue is potentially due to how the data is used in learning rather than the data itself.





How does this align with past results? \textsc{ECBD-easy} (Figure~\ref{fig:tradeoff-ecbd-easy}) shows a much more optimistic picture; recall that this is similar to the setting from \citet{mend}. In this case, MEND and fine-tuning both achieve results reasonably close to the train-on-test upper bound, with configurations that improve perplexity substantially with only mild specificity degradation. \textbf{Methods that succeed on injection of exact facts (e.g., injecting $y$ and reproducing it later) do not necessarily transfer to success in realistic knowledge propagation settings like ECBD.}


Finally, we plot the accuracy--specificity tradeoff curves computed on \textsc{Entity Inferences} (Figure~\ref{fig:tradeoff-entity-inferences}).
Table~\ref{tab:dataset_statistics} shows that the definition sentences of this dataset may contain the gold spans of the probe sentences but not always, making it between ECBD and \textsc{ECBD-easy} in this regard. Specificity numbers are less monotonic here than on ECBD, but we again see the trend of train-on-test quickly saturating accuracy. Like \textsc{ECBD-easy}, fine-tuning can lead to improvements on accuracy, in this case matching the performance of Prepend-Def. However, there remains a substantial gap with the gold setting, implying that \textbf{there are a certain number of examples that are not easily learnable by the current data setup.}

\begin{figure}
    \centering
    \begin{minipage}[t]{\linewidth}
    \begin{subfigure}{.475\linewidth}
    \centering
    \includegraphics[width=0.9\linewidth]{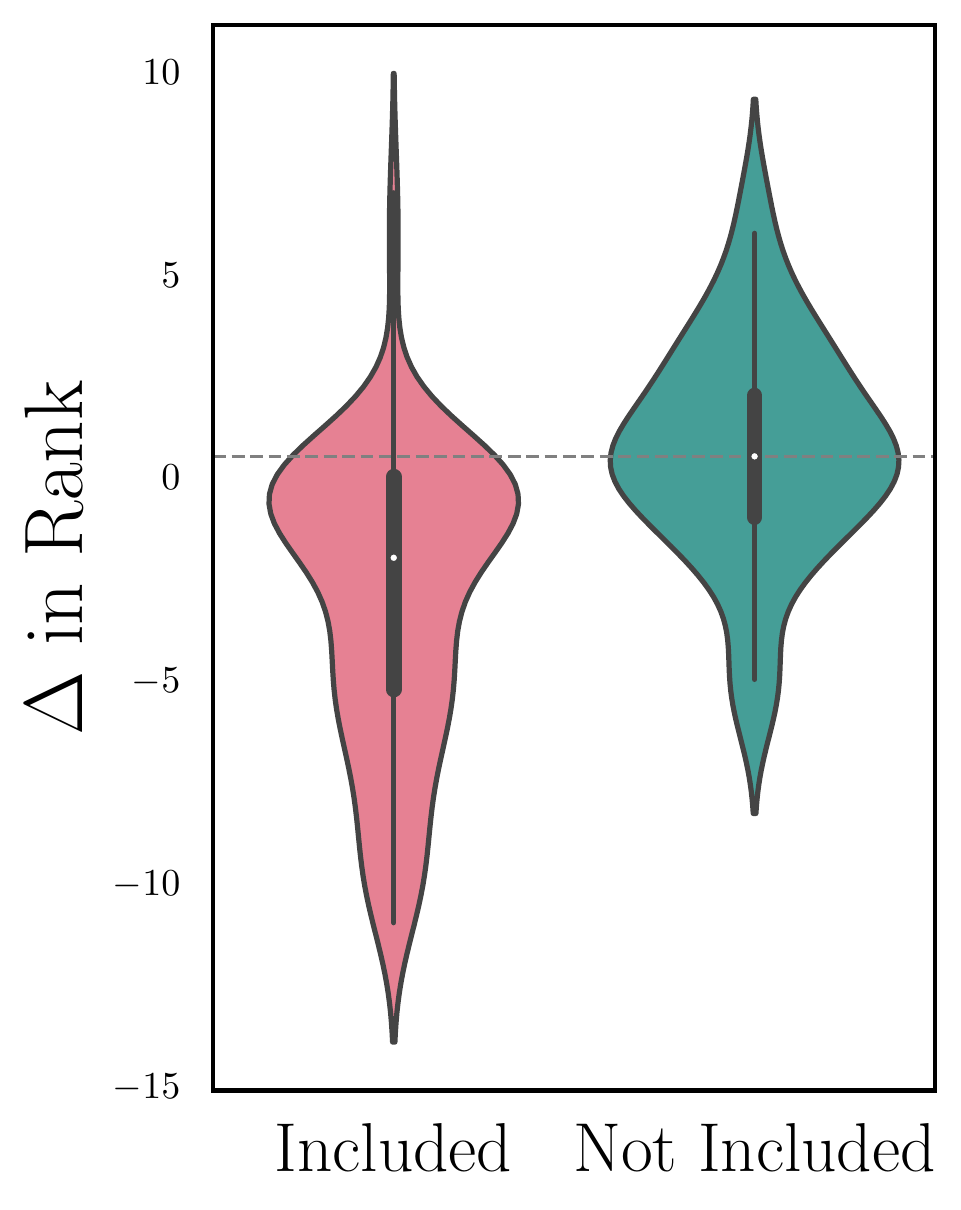}
    \end{subfigure}
    \begin{subfigure}{.475\linewidth}
    \centering
    \includegraphics[width=0.9\linewidth]{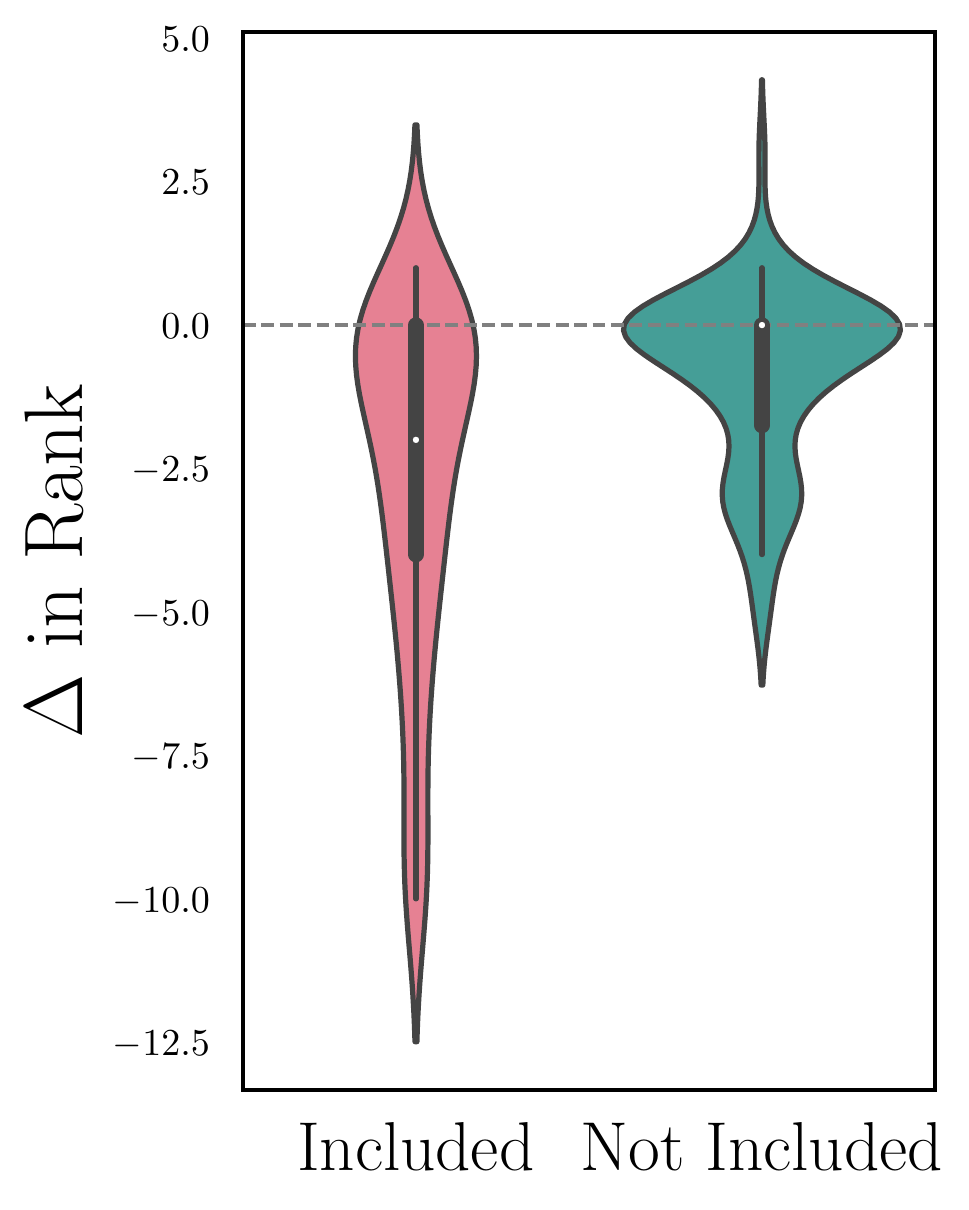}
    \end{subfigure}
    \subcaption{\textsc{Entity Inferences}}
    \label{fig:span_in_def_ei}
    \end{minipage}
    
    \begin{minipage}[t]{\linewidth}
    \centering
    \begin{subfigure}{.475\linewidth}
    \centering
    \includegraphics[width=0.9\linewidth]{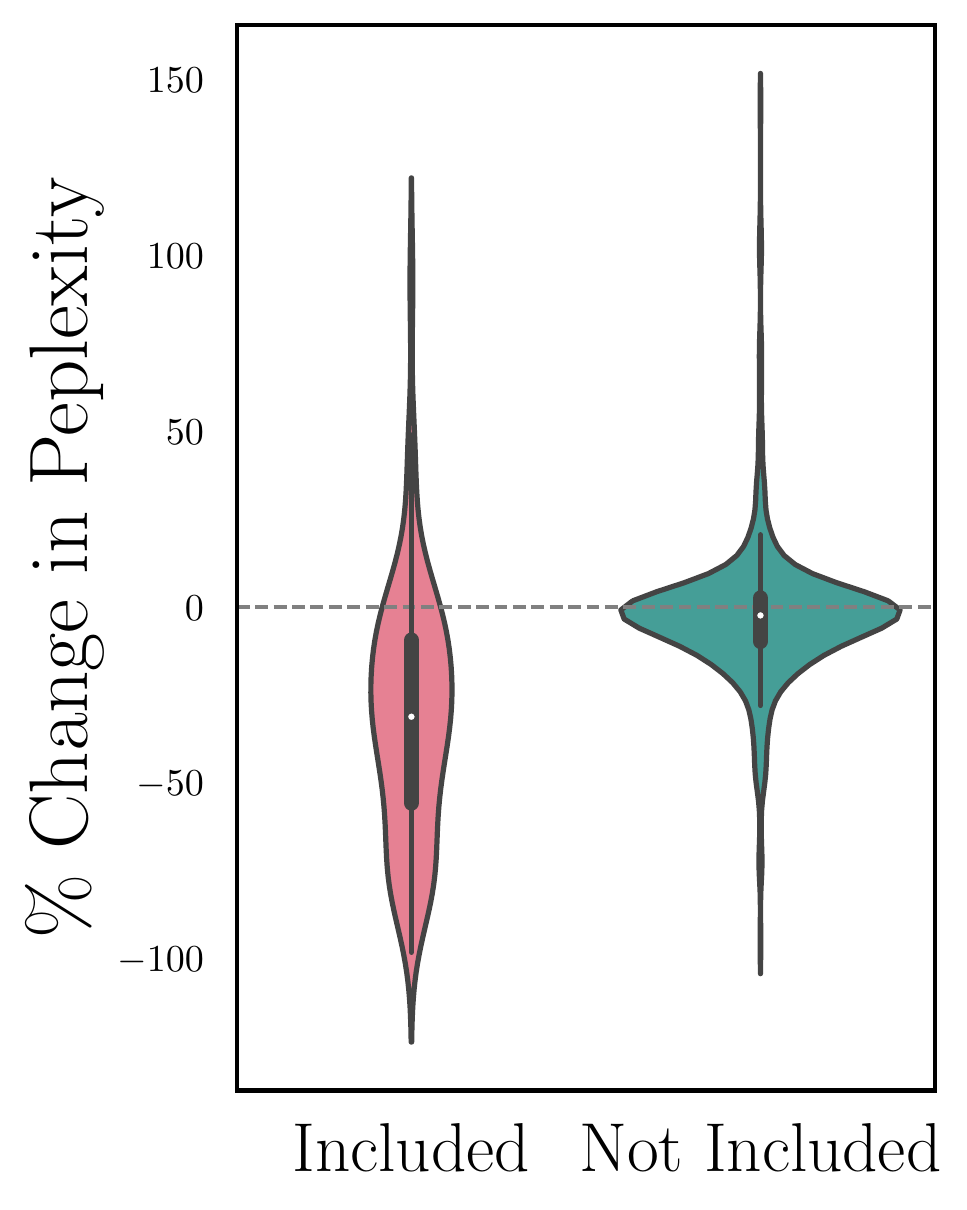}
    \end{subfigure}
    \begin{subfigure}{.475\linewidth}
    \centering
    \includegraphics[width=0.9\linewidth]{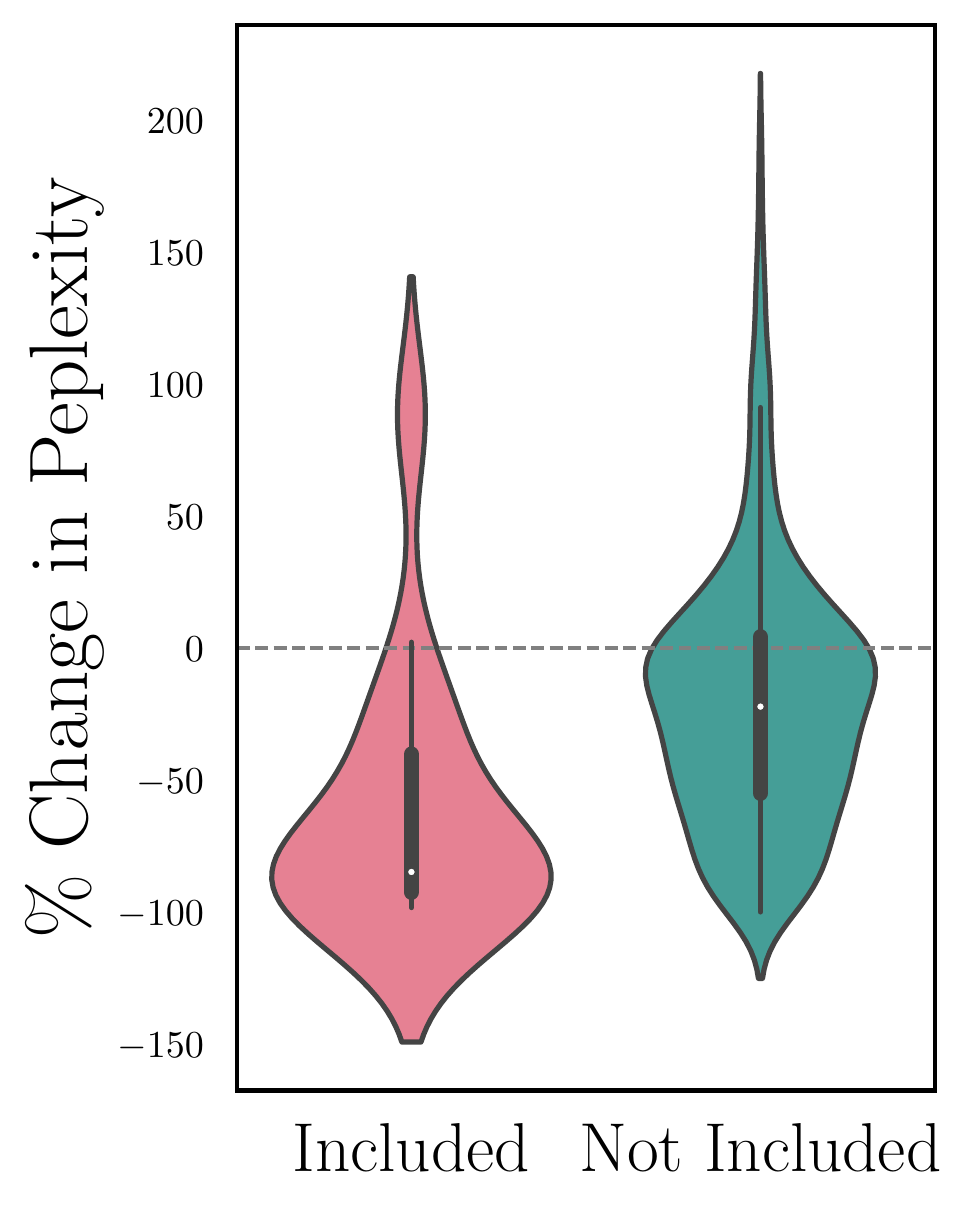}
    \end{subfigure}
    \subcaption{\textsc{ECBD}}
    \label{fig:span_in_def_ecbd}
    \end{minipage}

    \caption{(Left) Fine-tuning performance on GPT-Neo split by whether the gold span is included in the definition sentence or not. (Right) Input augmentation performance on GPT-Neo split by whether the gold span is included in the definition sentence or not.}
    \label{fig:span_in_def}
    
\end{figure}



\begin{figure}[t]
    \centering
    \begin{subfigure}{\linewidth}
    \centering
    \includegraphics[width=0.8\linewidth]{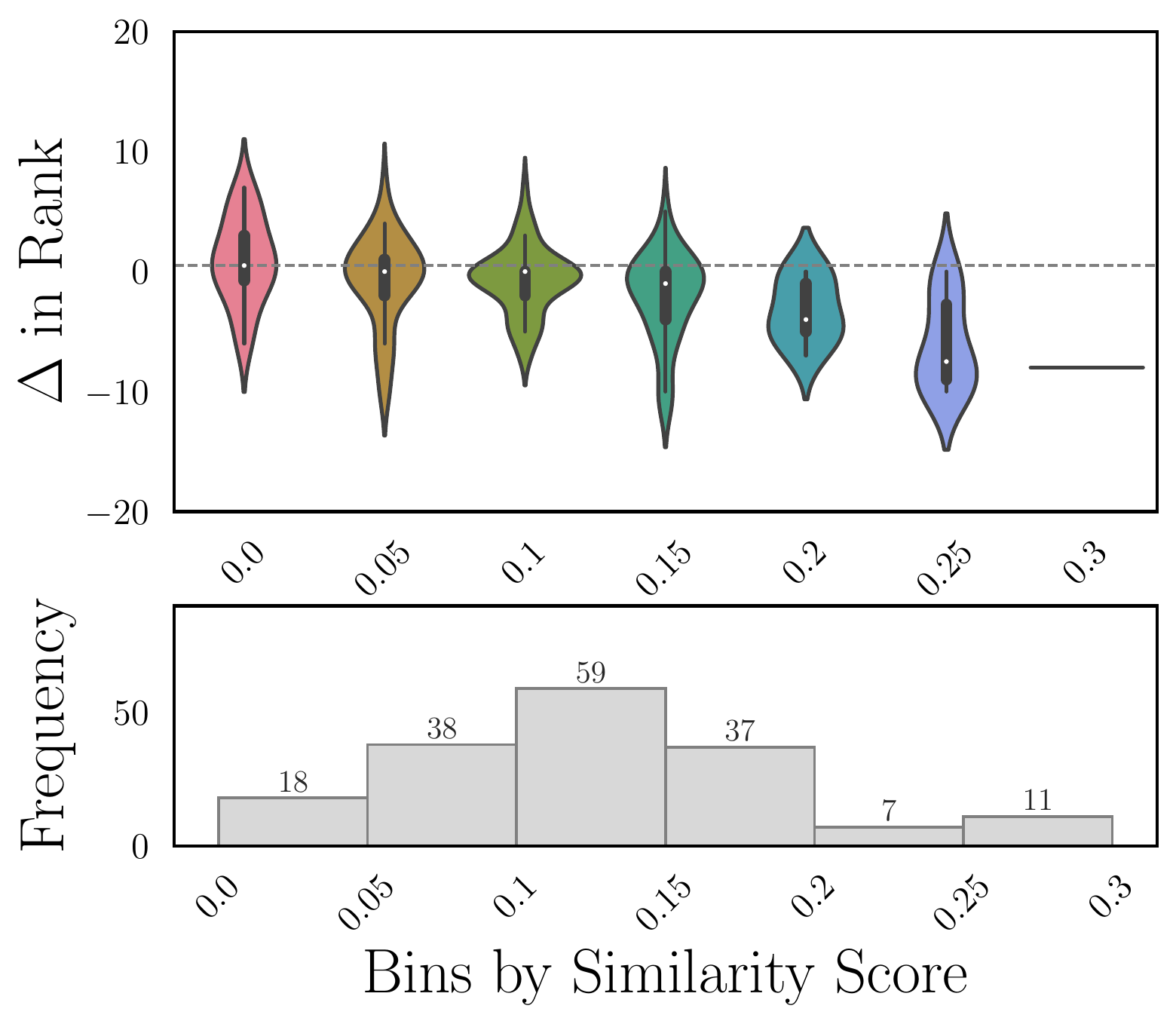}
    \caption{\textsc{Entity Inferences}}
    \end{subfigure}
    \begin{subfigure}{\linewidth}
    \centering
    \includegraphics[width=0.8\linewidth]{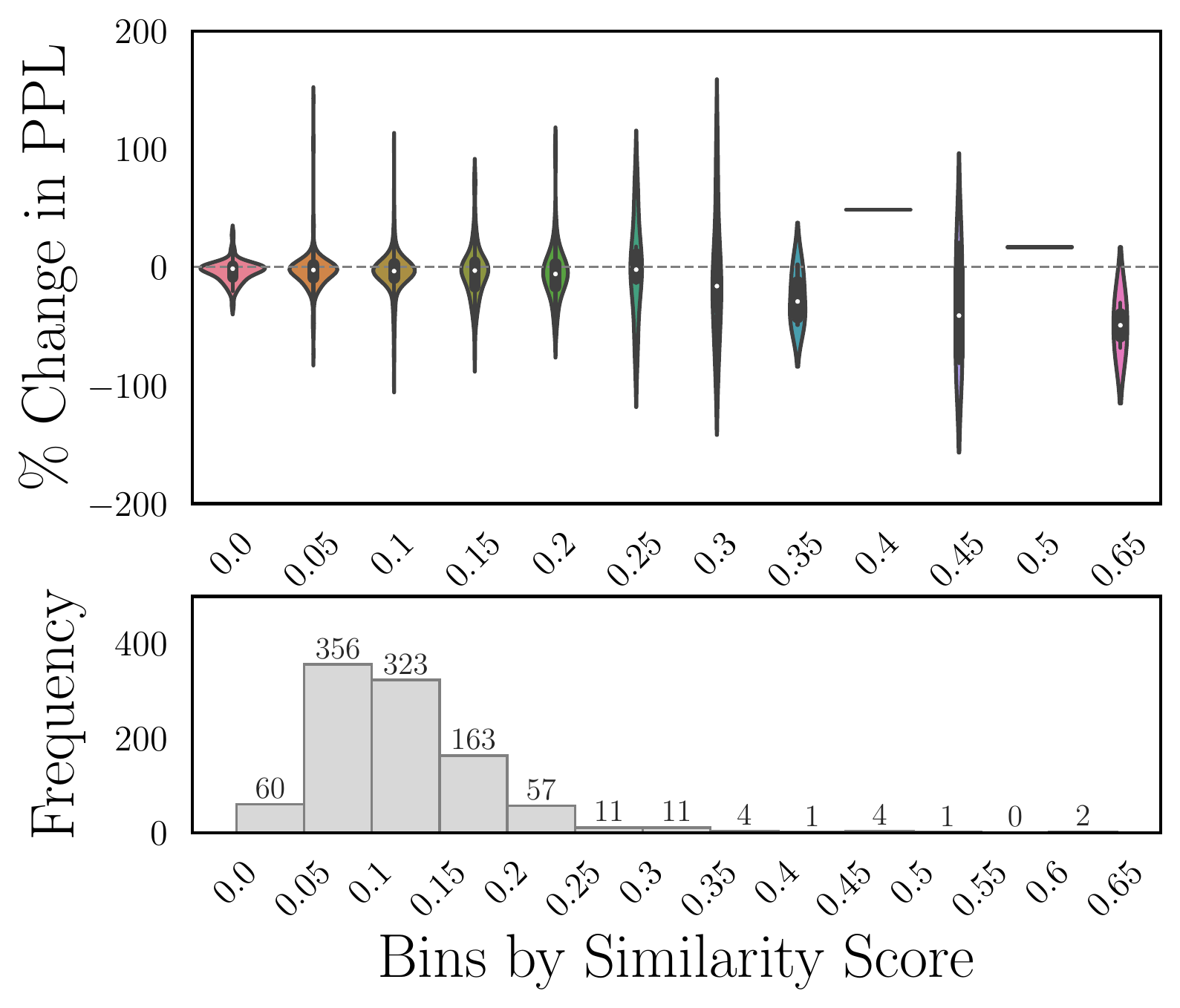}
    \caption{\textsc{ECBD}}
    \end{subfigure}\vspace{-5pt}
    \caption{Performance breakdown based on the lexical similarity (Jaccard similarity) between probe sentence $x_e$ and definition sentence $d_e$.}
    \label{fig:lexical-overlap-jaccard}
\end{figure}


\subsection{Information Overlap}\label{subsec:info_overlap}

\paragraph{Lexical overlap} We now examine the importance of overlap between the definition and the target span more closely. First, we look at instance-level behavior on our datasets stratified by whether the gold span is included in the definition or not. We select 92 such ``\emph{Included}'' examples in \textsc{Entity Inferences} and 152 from \textsc{ECBD-easy} and analyze the delta in the rank of the gold label and percent change in perplexity respectively.


Figure~\ref{fig:span_in_def_ei} shows violin plots of the performance gaps within the two groups. In both datasets, the performance improves on average (plot mean below 0) when the gold spans are included in the definition sentences, suggesting that \textbf{the lexical overlap between the definition and probe sentences correlates with the model performance}.
This trend on \textsc{ECBD} is even stronger with input augmentation (Figure~\ref{fig:span_in_def_ecbd}).
However, the majority of \textsc{ECBD} probe sentences fall into the \emph{Not Included} category, and we see here that very few examples in this category have substantial perplexity improvements, most having small changes around zero. 

\textsc{Entity Inferences} shows a slightly more optimistic picture for \emph{Not Included} cases. 




\paragraph{Soft overlap} Although direct inclusion of the answer span is clearly valuable, do we see any improvements when there is \emph{soft overlap} between the definition and target span; that is, the content may be similar even if not exactly the same?

We investigate the information overlap using both lexical (e.g., Jaccard similarity, Rouge) and semantic (e.g, BERTScore~\cite{bert-score}) similarity measurements between the probe sentence and the definition sentence. For each dataset, we divide the examples into bins based on the similarity scores and report the performance differences between the base model and the fine-tuned model per bin (change in rank of the gold answer on \textsc{Entity Inferences} and perplexity change on \textsc{ECBD}). 

Figure~\ref{fig:lexical-overlap-jaccard} shows violin plots of the performance gaps within each bin constructed using Jaccard similarity (a larger value mean the definition and probe sentences are similar).
For \textsc{Entity Inferences}, we observe that the bins with larger similarity scores have progressively more negative $\Delta$ in rank.
Surprisingly, we do not see a similar trend for ECBD. Not only is it the case that there are fewer examples in ECBD exhibiting high overlap, but among the distribution of examples that is present, there is almost no perceptible correlation between the amount of overlap and the percentage change in perplexity. This suggests that not only is the data distribution in ECBD different, but \textbf{the nature of the inferences themselves can be qualitatively different and more challenging}. We believe this further underscores that new techniques are needed to handle knowledge propagation in the real world.





\section{Related Work}

\paragraph{Keeping Language Models Up to Date} One line of recent work have explored the development and evaluation of language models that are updated over time \cite{jang-etal-2022-temporalwiki}. While ECBD~\cite{onoe-etal-2022-entity} focuses solely on evaluating knowledge of new entities, several benchmarks have been proposed for evaluating facts about existing entities that have changed over time as open-retrieval~\cite{zhang-choi-2021-situatedqa} or cloze-style~\cite{dhingra-etal-2022-time} question answering. Other work has found success in keeping LMs up-to-date by continuing pretraining~\cite{jin-etal-2022-lifelong-pretraining} and applying domain adaptation techniques~\cite{jang-etal-2022-improving}. Beyond these and the editing approaches we have discussed previously, a line of work has looked at identifying a small, localized set of weights that are responsible for reflecting the memorized fact~\cite{geva2020transformer} and editing only that small set of parameters~\cite{rome,Dai2021KnowledgeNI}. Finally, \citet{choi-etal-2022-promptinjection} also contrast prepending information with fine-tuning and find that fine-tuning generally works worse, framing their approach as distillation.

\paragraph{Content Transfer and Knowledge Acquisition}
The tasks and setting we explore in our work are closely related to that of \citet{Peter_West_2022}, which explores whether LMs can generate statements about an entity that are consistent with a provided description of that entity. However, they do not explore updating model parameters from these descriptions. 

\section{Conclusion}

In this work, we explored the \emph{entity knowledge propagation} setting: to what extent can descriptions of new entities be injected into language models? We find that while fine-tuning models or using efficient update strategies enables models to reproduce exact facts from descriptions, performing inferences based on those facts is substantially harder. We characterize several approaches on two datasets and conclude that update strategies lag the performance of simply prepending the definition in the context, suggesting that more work is needed.

\section*{Limitations}


Entity knowledge propagation focuses on updating LMs' knowledge about emerging entities.
However, there might be cases where knowledge about existing entities needs to be updated (e.g., regime change, new champion, and renaming etc.).
We intentionally exclude these cases since they can easily become intractable due to their complexity.
For example, an organization changing its name could theoretically reflect a large number of entities that have relations to that organization. 
By investigating model behavior when a LM encounters new information which is completely unseen during pretraining, we can experiment in a controlled environment. We find ample challenges unaddressed by current research even in this setting.


Our experiments are conducted on English language models only. While we believe the results can generalize to multilingual models, it is conceivable that the internal representations of these models make them more or less amenable to the sorts of updating explored here. More work is needed to benchmark these techniques in broader settings such as with larger language models and newer parameter-tuning approaches. 

\section*{Acknowledgments}

This work was partially supported by NSF Grant IIS-1814522, NSF CAREER Award IIS-2145280, a grant from Open Philanthropy, and by the Air Force Research Laboratory (AFRL), DARPA for the KAIROS program under agreement number FA8750-19-2-1003. The views and conclusions contained herein are those of the authors and should not be interpreted as necessarily representing the official policies, either expressed or implied, of DARPA, or the U.S.~Government. The U.S.~Government is authorized to reproduce and distribute reprints for governmental purposes notwithstanding any copyright annotation therein.

\bibliography{anthology,custom}

\begin{thebibliography}{25}
\expandafter\ifx\csname natexlab\endcsname\relax\def\natexlab#1{#1}\fi

\bibitem[{Black et~al.(2021)Black, Gao, Wang, Leahy, and Biderman}]{gpt-neo}
Sid Black, Leo Gao, Phil Wang, Connor Leahy, and Stella Biderman. 2021.
\newblock \href {https://doi.org/10.5281/zenodo.5297715} {{GPT-Neo: Large Scale
  Autoregressive Language Modeling with Mesh-Tensorflow}}.

\bibitem[{Choi et~al.(2022)Choi, Jo, Jang, and
  Seo}]{choi-etal-2022-promptinjection}
Eunbi Choi, Yongrae Jo, Joel Jang, and Minjoon Seo. 2022.
\newblock \href {https://arxiv.org/abs/2206.11349} {{Prompt Injection:
  Parameterization of Fixed Inputs}}.
\newblock \emph{arXiv}, abs/2206.11349.

\bibitem[{Dai et~al.(2021)Dai, Dong, Hao, Sui, and Wei}]{Dai2021KnowledgeNI}
Damai Dai, Li~Dong, Yaru Hao, Zhifang Sui, and Furu Wei. 2021.
\newblock \href {https://arxiv.org/abs/2104.08696} {Knowledge neurons in
  pretrained transformers}.
\newblock \emph{arXiv}, abs/2104.08696.

\bibitem[{De~Cao et~al.(2021)De~Cao, Aziz, and Titov}]{Nicola_De_Cao_21_KE}
Nicola De~Cao, Wilker Aziz, and Ivan Titov. 2021.
\newblock \href {https://doi.org/10.18653/v1/2021.emnlp-main.522} {Editing
  factual knowledge in language models}.
\newblock In \emph{Proceedings of the 2021 Conference on Empirical Methods in
  Natural Language Processing}, pages 6491--6506, Online and Punta Cana,
  Dominican Republic. Association for Computational Linguistics.

\bibitem[{Dhingra et~al.(2022{\natexlab{a}})Dhingra, Cole, Eisenschlos,
  Gillick, Eisenstein, and Cohen}]{Bhuwan_Dhingra_21}
Bhuwan Dhingra, Jeremy~R. Cole, Julian~Martin Eisenschlos, Daniel Gillick,
  Jacob Eisenstein, and William~W. Cohen. 2022{\natexlab{a}}.
\newblock \href {https://doi.org/10.1162/tacl_a_00459} {Time-aware language
  models as temporal knowledge bases}.
\newblock volume~10, pages 257--273, Cambridge, MA. MIT Press.

\bibitem[{Dhingra et~al.(2022{\natexlab{b}})Dhingra, Cole, Eisenschlos,
  Gillick, Eisenstein, and Cohen}]{dhingra-etal-2022-time}
Bhuwan Dhingra, Jeremy~R. Cole, Julian~Martin Eisenschlos, Daniel Gillick,
  Jacob Eisenstein, and William~W. Cohen. 2022{\natexlab{b}}.
\newblock \href {https://doi.org/10.1162/tacl_a_00459} {Time-aware language
  models as temporal knowledge bases}.
\newblock \emph{Transactions of the Association for Computational Linguistics},
  10:257--273.

\bibitem[{Gao et~al.(2020)Gao, Biderman, Black, Golding, Hoppe, Foster, Phang,
  He, Thite, Nabeshima, Presser, and Leahy}]{pile}
Leo Gao, Stella Biderman, Sid Black, Laurence Golding, Travis Hoppe, Charles
  Foster, Jason Phang, Horace He, Anish Thite, Noa Nabeshima, Shawn Presser,
  and Connor Leahy. 2020.
\newblock The {P}ile: An 800gb dataset of diverse text for language modeling.
\newblock \emph{arXiv preprint arXiv:2101.00027}.

\bibitem[{Geva et~al.(2021)Geva, Schuster, Berant, and
  Levy}]{geva2020transformer}
Mor Geva, Roei Schuster, Jonathan Berant, and Omer Levy. 2021.
\newblock \href {https://doi.org/10.18653/v1/2021.emnlp-main.446} {Transformer
  feed-forward layers are key-value memories}.
\newblock In \emph{Proceedings of the 2021 Conference on Empirical Methods in
  Natural Language Processing}, pages 5484--5495, Online and Punta Cana,
  Dominican Republic. Association for Computational Linguistics.

\bibitem[{Gururangan et~al.(2020)Gururangan, Marasovi{\'c}, Swayamdipta, Lo,
  Beltagy, Downey, and Smith}]{gururangan-etal-2020-dont}
Suchin Gururangan, Ana Marasovi{\'c}, Swabha Swayamdipta, Kyle Lo, Iz~Beltagy,
  Doug Downey, and Noah~A. Smith. 2020.
\newblock \href {https://doi.org/10.18653/v1/2020.acl-main.740} {Don{'}t stop
  pretraining: Adapt language models to domains and tasks}.
\newblock In \emph{Proceedings of the 58th Annual Meeting of the Association
  for Computational Linguistics}, pages 8342--8360, Online. Association for
  Computational Linguistics.

\bibitem[{Jang et~al.(2022{\natexlab{a}})Jang, Ye, Lee, Yang, Shin, Han, Kim,
  and Seo}]{jang-etal-2022-temporalwiki}
Joel Jang, Seonghyeon Ye, Changho Lee, Sohee Yang, Joongbo Shin, Janghoon Han,
  Gyeonghun Kim, and Minjoon Seo. 2022{\natexlab{a}}.
\newblock \href {https://arxiv.org/abs/2204.14211} {{TemporalWiki: A Lifelong
  Benchmark for Training and Evaluating Ever-Evolving Language Models}}.
\newblock In \emph{Proceedings of the 2022 Conference on Empirical Methods in
  Natural Language Processing}.

\bibitem[{Jang et~al.(2022{\natexlab{b}})Jang, Ye, Yang, Shin, Han, Kim, Choi,
  and Seo}]{jang-etal-2022-continual}
Joel Jang, Seonghyeon Ye, Sohee Yang, Joongbo Shin, Janghoon Han, Gyeonghun
  Kim, Stanley~Jungkyu Choi, and Minjoon Seo. 2022{\natexlab{b}}.
\newblock \href {https://arxiv.org/abs/2110.03215} {{Towards Continual
  Knowledge Learning of Language Models}}.
\newblock In \emph{Proceedings of the International Conference on Learning
  Representations (ICLR)}.

\bibitem[{Jang et~al.(2022{\natexlab{c}})Jang, Lee, Park, Kang, Lee, Bae, and
  Jung}]{jang-etal-2022-improving}
Yunah Jang, Dongryeol Lee, Hyung~Joo Park, Taegwan Kang, Hwanhee Lee, Hyunkyung
  Bae, and Kyomin Jung. 2022{\natexlab{c}}.
\newblock \href {https://doi.org/10.18653/v1/2022.dialdoc-1.15} {Improving
  multiple documents grounded goal-oriented dialog systems via diverse
  knowledge enhanced pretrained language model}.
\newblock In \emph{Proceedings of the Second DialDoc Workshop on
  Document-grounded Dialogue and Conversational Question Answering}, pages
  136--141, Dublin, Ireland. Association for Computational Linguistics.

\bibitem[{Jin et~al.(2022)Jin, Zhang, Zhu, Xiao, Li, Wei, Arnold, and
  Ren}]{jin-etal-2022-lifelong-pretraining}
Xisen Jin, Dejiao Zhang, Henghui Zhu, Wei Xiao, Shang-Wen Li, Xiaokai Wei,
  Andrew Arnold, and Xiang Ren. 2022.
\newblock \href {https://doi.org/10.18653/v1/2022.naacl-main.351} {Lifelong
  pretraining: Continually adapting language models to emerging corpora}.
\newblock In \emph{Proceedings of the 2022 Conference of the North American
  Chapter of the Association for Computational Linguistics: Human Language
  Technologies}, pages 4764--4780, Seattle, United States. Association for
  Computational Linguistics.

\bibitem[{Lazaridou et~al.(2021)Lazaridou, Kuncoro, Gribovskaya, Agrawal,
  Liska, Terzi, Gimenez, de~Masson~d'Autume, Kocisky, Ruder, Yogatama, Cao,
  Young, and Blunsom}]{Angeliki_Lazaridou_21}
Angeliki Lazaridou, Adhiguna Kuncoro, Elena Gribovskaya, Devang Agrawal, Adam
  Liska, Tayfun Terzi, Mai Gimenez, Cyprien de~Masson~d'Autume, Tomas Kocisky,
  Sebastian Ruder, Dani Yogatama, Kris Cao, Susannah Young, and Phil Blunsom.
  2021.
\newblock {Mind the Gap: Assessing Temporal Generalization in Neural Language
  Models}.
\newblock In \emph{Advances in Neural Information Processing Systems
  (NeurIPS)}.

\bibitem[{Meng et~al.(2022)Meng, Bau, Andonian, and Belinkov}]{rome}
Kevin Meng, David Bau, Alex Andonian, and Yonatan Belinkov. 2022.
\newblock {Locating and Editing Factual Associations in GPT}.
\newblock In \emph{Advances in Neural Information Processing Systems
  (NeurIPS)}.

\bibitem[{Mitchell et~al.(2022)Mitchell, Lin, Bosselut, Finn, and
  Manning}]{mend}
Eric Mitchell, Charles Lin, Antoine Bosselut, Chelsea Finn, and Christopher~D
  Manning. 2022.
\newblock {Fast Model Editing at Scale}.
\newblock In \emph{International Conference on Learning Representations
  (ICLR)}.

\bibitem[{Onoe et~al.(2022)Onoe, Zhang, Choi, and
  Durrett}]{onoe-etal-2022-entity}
Yasumasa Onoe, Michael Zhang, Eunsol Choi, and Greg Durrett. 2022.
\newblock \href {https://doi.org/10.18653/v1/2022.findings-naacl.52} {Entity
  cloze by date: What {LM}s know about unseen entities}.
\newblock In \emph{Findings of the Association for Computational Linguistics:
  NAACL 2022}, pages 693--702, Seattle, United States. Association for
  Computational Linguistics.

\bibitem[{Radford et~al.(2019)Radford, Wu, Child, Luan, Amodei, and
  Sutskever}]{radford2019language}
Alec Radford, Jeff Wu, Rewon Child, David Luan, Dario Amodei, and Ilya
  Sutskever. 2019.
\newblock {Language Models are Unsupervised Multitask Learners}.

\bibitem[{Raffel et~al.(2020)Raffel, Shazeer, Roberts, Lee, Narang, Matena,
  Zhou, Li, and Liu}]{T5}
Colin Raffel, Noam Shazeer, Adam Roberts, Katherine Lee, Sharan Narang, Michael
  Matena, Yanqi Zhou, Wei Li, and Peter~J. Liu. 2020.
\newblock \href {http://jmlr.org/papers/v21/20-074.html} {{Exploring the Limits
  of Transfer Learning with a Unified Text-to-Text Transformer}}.
\newblock \emph{Journal of Machine Learning Research}, 21(140):1--67.

\bibitem[{Sinitsin et~al.(2020)Sinitsin, Plokhotnyuk, Pyrkin, Popov, and
  Babenko}]{Anton_Sinitsin_20}
Anton Sinitsin, Vsevolod Plokhotnyuk, Dmitriy Pyrkin, Sergei Popov, and Artem
  Babenko. 2020.
\newblock {Editable Neural Networks}.
\newblock In \emph{International Conference on Learning Representations
  (ICLR)}.

\bibitem[{West et~al.(2022)West, Quirk, Galley, and Choi}]{Peter_West_2022}
Peter West, Chris Quirk, Michel Galley, and Yejin Choi. 2022.
\newblock \href {https://doi.org/10.18653/v1/2022.findings-acl.294} {Probing
  factually grounded content transfer with factual ablation}.
\newblock In \emph{Findings of the Association for Computational Linguistics:
  ACL 2022}, pages 3732--3746, Dublin, Ireland. Association for Computational
  Linguistics.

\bibitem[{Wolf et~al.(2020)Wolf, Debut, Sanh, Chaumond, Delangue, Moi, Cistac,
  Rault, Louf, Funtowicz, Davison, Shleifer, von Platen, Ma, Jernite, Plu, Xu,
  Le~Scao, Gugger, Drame, Lhoest, and Rush}]{wolf-etal-2020-transformers}
Thomas Wolf, Lysandre Debut, Victor Sanh, Julien Chaumond, Clement Delangue,
  Anthony Moi, Pierric Cistac, Tim Rault, Remi Louf, Morgan Funtowicz, Joe
  Davison, Sam Shleifer, Patrick von Platen, Clara Ma, Yacine Jernite, Julien
  Plu, Canwen Xu, Teven Le~Scao, Sylvain Gugger, Mariama Drame, Quentin Lhoest,
  and Alexander Rush. 2020.
\newblock \href {https://doi.org/10.18653/v1/2020.emnlp-demos.6} {Transformers:
  State-of-the-art natural language processing}.
\newblock In \emph{Proceedings of the 2020 Conference on Empirical Methods in
  Natural Language Processing: System Demonstrations}, pages 38--45, Online.
  Association for Computational Linguistics.

\bibitem[{Zhang and Choi(2021)}]{zhang-choi-2021-situatedqa}
Michael Zhang and Eunsol Choi. 2021.
\newblock \href {https://doi.org/10.18653/v1/2021.emnlp-main.586}
  {{S}ituated{QA}: Incorporating extra-linguistic contexts into {QA}}.
\newblock In \emph{Proceedings of the 2021 Conference on Empirical Methods in
  Natural Language Processing}, pages 7371--7387, Online and Punta Cana,
  Dominican Republic. Association for Computational Linguistics.

\bibitem[{Zhang et~al.(2020)Zhang, Kishore, Wu, Weinberger, and
  Artzi}]{bert-score}
Tianyi Zhang, Varsha Kishore, Felix Wu, Kilian~Q. Weinberger, and Yoav Artzi.
  2020.
\newblock {BERTScore: Evaluating Text Generation with BERT}.
\newblock In \emph{International Conference on Learning Representations
  (ICLR)}.

\bibitem[{Zhu et~al.(2020)Zhu, Rawat, Zaheer, Bhojanapalli, Li, Yu, and
  Kumar}]{Chen_Zhu_2020}
Chen Zhu, Ankit~Singh Rawat, Manzil Zaheer, Srinadh Bhojanapalli, Daliang Li,
  Felix Yu, and Sanjiv Kumar. 2020.
\newblock \href {https://arxiv.org/abs/2012.00363} {Modifying memories in
  transformer models}.
\newblock \emph{arXiv}, abs/2012.00363.

\end{thebibliography}
\bibliographystyle{acl_natbib}

\newpage
\appendix
\section{Appendix}

\subsection{Licensing}

T5 is released under the Apache v2.0 license. GPT-2 and GPT-Neo is released under the MIT license. Wikipedia and ECBD are both licensed under CC BY-SA.

\subsection{Harmful Data Instances}

In creating our dataset of entity inferences, we, the authors, inspect and only create examples that do not contain offensive or harmful content. All other data used is publically availible from Wikipedia. Experiments and data are all in English.

\subsection{Modeling Details}\label{app:hyperparams}
The main hyperparameters were the size of the training batch (always 1), the size of the validation batch (always 1), the number of epochs for training (in the finetuning case), and the learning rate. The number of training epochs was 5 for ECBD experiments and 10 for Entity Inferences experiments, and the learning rate was 3e-6 on ECBD and 5e-4 on Entity Inferences. 

We run all experiments on a machine with four Quadro RTX 8000 GPUs for less than 4 GPU hours. All experiments and results reflect just a single run. We use the Huggingface Transformers packages \cite{wolf-etal-2020-transformers} for running our models and analysis.

\subsection{\textsc{Entity Inferences}}\label{app:entit_inference_data_constuction}

\paragraph{Data Construction Details} We first curate entities tagged with TV shows and natural disasters from English Wikipedia and their definition sentences from the 2020 and 2021 subsets of ECBD. In addition to real entities, we generate examples of "fake" person where we fabricate person names along with their definitions (e.g., Leighanna Smith (born July 21, 1970) is an American film director, screenwriter, and producer...). 

Then, we manually craft two types of probe sentences: explicit and implicit. The explicit probe sentences ask information that is explicitly stated in the definition sentence (e.g., genre of a TV show). On the other hand, the implicit probe sentences require commonsense-like information (e.g., people watch a TV show, don't eat a TV show.).



For each entity, we manually write several types of probe sentences that test LMs' knowledge in different ways. The \emph{explicit} probe sentences ask about information that are explicitly stated in the definition sentence (e.g., genre of a TV show, occupation of a person). On the other hand, the \emph{implicit} probe sentences require commonsense-like information (e.g., people watch a TV show, don't eat a TV show.).
Finally, we write answer candidates (between 6 to 12) for each type of probe sentences. On average, one example has 10 answer candidates.  
Each example consists of elements listed below (example in Table~\ref{tab:app_examples2}).


\subsection{More Similarity Scores}\label{app:similarity}
\begin{figure*}
    \centering
    
    \begin{subfigure}[H]{\linewidth}
    \centering
      \includegraphics[width=\textwidth]{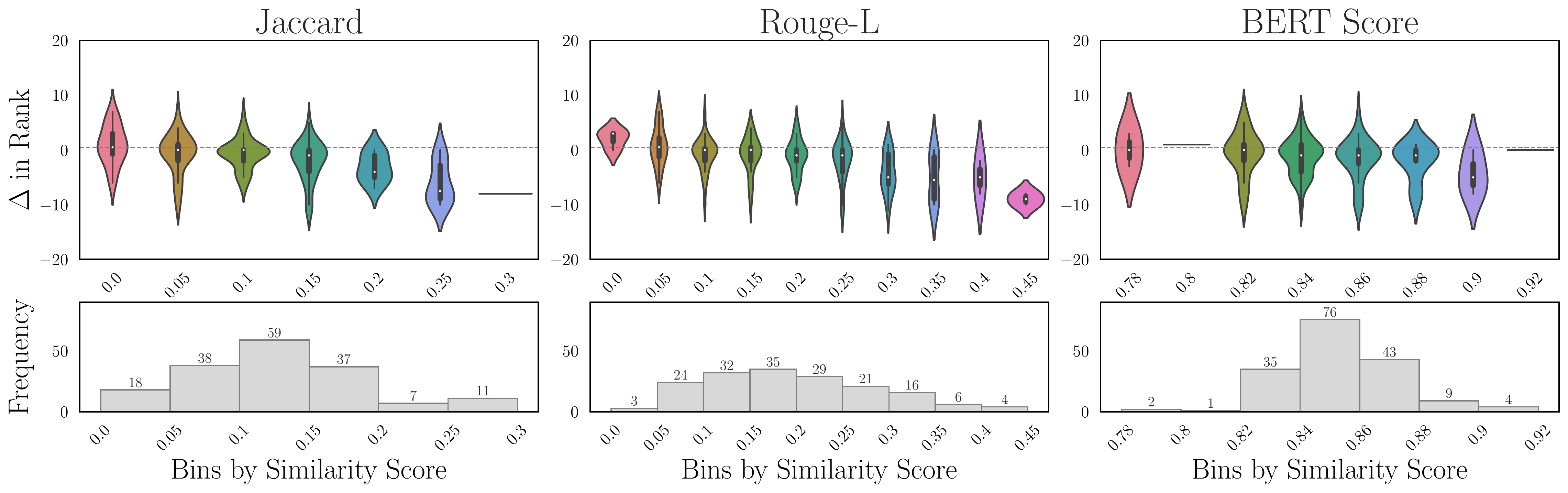}
      \caption{\textsc{Entity Inferences}}
    \end{subfigure}\\
    \vspace{4pt}
    \begin{subfigure}[H]{\linewidth}
    \centering
      \includegraphics[width=\textwidth]{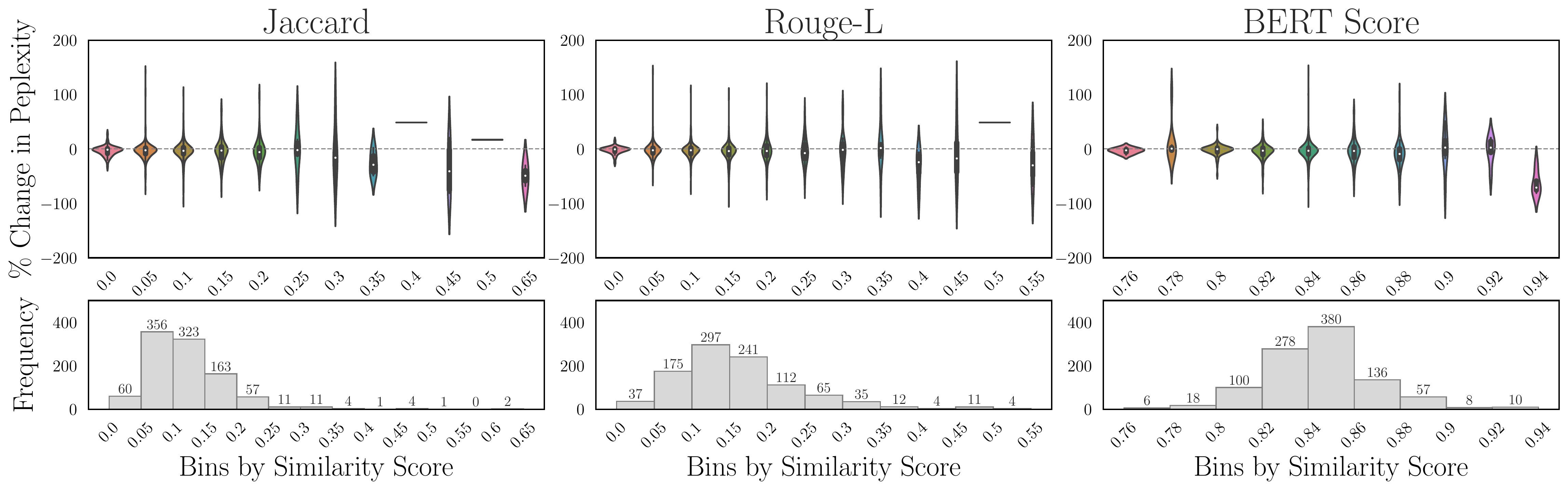}
      \caption{\textsc{ECBD}}
    \end{subfigure}
    \caption{Performance breakdown based on the lexical similarity between probe sentence $x_e$ and definition sentence $d_e$.}
    \label{fig:lexical-overlap}
\end{figure*}

Figure~\ref{fig:lexical-overlap} compares two lexical (Jaccard and Rouge-L) and one semantic (BERT Score) similarity scores.

\subsection{Analysis of ROME}

\subsubsection{Comparison of datasets}
The Counterfactual dataset was one of the datasets created and used by \citep{rome}. It consisted of a set of "counterfacts" - facts that are altered slightly. For example, one entry in this dataset is "The Eiffel Tower is located in the City of Rome". 

As one can see in Table 4, the three datasets scale in complexity. Counterfactual usually includes known entities (subjects) and known labels (objects). Entity Inferences usually contains unknown entities, but its labels are often known. Lastly, ECBD not only has unknown entities, but it also sometimes contains non-descriptive labels. This may explain why it obtained such drastic increases in perplexity on ECBD.

\renewcommand{\arraystretch}{1}
\begin{table*}[t]
	\centering
	\footnotesize
	\setlength{\tabcolsep}{4pt}
	\def\arraystretch{1.25}
	\begin{tabular}{ p{1.5cm}p{5cm}p{5cm}p{3cm}}
		\toprule
    		%
  {\sc Entity} & {{\sc Definition}} & {{\sc Probe Sentences}} & {{\sc Gold Label}} \\\midrule
	     { 2020 Vuelta a España} & { The 2020 Vuelta a España was the 75th edition of the Vuelta a España, one of cycling's three grand tours.}  & {The full route of the 2020 Vuelta a España was announced on <MASK> in Madrid.} & {Tuesday 17 December 2019}  \\ \midrule
		 M1 & {The Apple M1 is an ARM-based system on a chip (SoC).} & {The M1 contains <MASK> in a 16-core Neural Engine, capable of executing 11 trillion operations per second.} & {dedicated neural network hardware}   \\  \midrule
		 Dixie Fire & {The Dixie Fire is an active wildfire in Butte, Plumas, Lassen, and Tehama Counties, California.}  & {Smoke from the Dixie Fire caused <MASK> across the Western United States, including as far east of California as Utah and Colorado..} & {unhealthy air quality}   \\ \midrule
		{Cravity} &{Cravity () is a South Korean boy band formed by Starship Entertainment}  & {On August 13, at the 2020 Soribada Awards, Cravity won the "New Artist Award", <MASK> since debut.} & {their first award} \\
		\bottomrule 
	\end{tabular}
 \vspace{-0.4em}
	\caption{Examples from ECBD.} \label{tab:app_examples1}
 \label{table:ecbd_examples}
\end{table*}

\renewcommand{\arraystretch}{1}
\begin{table*}[t]
	\centering
	\footnotesize
	\setlength{\tabcolsep}{4pt}
	\def\arraystretch{1.25}
	\begin{tabular}{ p{3cm}p{6cm}p{5cm}p{1cm}}
		\toprule
    		%
  {\sc Entity} & {{\sc Definition}} & {{\sc Probe Sentences}} & {{\sc Gold Label}} \\\midrule
	     {Cyclone Niran} & {Severe Tropical Cyclone Niran was a very powerful tropical cyclone that brought severe impacts to extreme Northeastern Australia and nearly made landfall in New Caledonia in February and March 2021.}  & {Cyclone Niran left widespread damage in <MASK>.} & {Australia}  \\ \midrule
		 2020 Lekki shooting & {On the night of 20 October 2020, at about 6:50p.m., members of the Nigerian Army opened fire on peaceful End SARS protesters at the Lekki toll gate in Lagos State, Nigeria} & {2020 Lekki shooting happened near my house, so my family and I <MASK> from the area.} & {escaped}   \\ \midrule
		 Ronald Deschamplains & {Roland Deschamplains (born September 21, 1989), better known by his stage name Desham, is an American singer , songwriter, and dancer who has sold over 30 million singles and has achieved eleven Platinum singles.}  & {Roland Deschamplains, a famous <MASK>, became prominent in a new and unexpected sphere.} & {singer}   \\ \midrule
		{The Great} &{The Great is a 2020 comedy-drama television series described by its commissioner Hulu as 'anti-historical' loosely based on the rise to power of Catherine the Great, Empress of All Russia.}  & {Some people think The Great is very <MASK>.} & {funny} \\
		\bottomrule 
	\end{tabular}
 \vspace{-0.4em}
	\caption{Examples from Entity Inferences} \label{tab:app_examples2}
 \label{table:entity_inference_examples}
\end{table*}

\begin{table*}[t]
	\centering
	\footnotesize
	\setlength{\tabcolsep}{4pt}
	\def\arraystretch{1.25}
	\begin{tabular}{p{3cm} p{11cm}}
		\toprule
 		 \multicolumn{1}{l}{Dataset} & \multicolumn{1}{l}{Example}    \\
		\midrule
	    {Counterfactual} & {"The \underline{Eiffel Tower} is located in the City of \textbf{Rome}"}   \\
      \midrule
      {Entity Inferences} & {"Severe Tropical \underline{Cyclone Niran} was a very powerful tropical cyclone that brought severe impacts to extreme Northeastern \textbf{Australia}"}  \\
      \midrule
      {ECBD} & { "\underline{Gamma} variant, also known as lineage P.1, \textbf{is one of the} variants of SARS-CoV-2, the virus that causes COVID-19."}   \\
		\bottomrule 

	\end{tabular}
	\caption{Comparison of one example of three datasets. The subject is underlined and the object is bolded.} \label{tab:app_examples3}
 \label{table:dataset_comparison}
\end{table*}
\begin{table*}[t]
	\centering
	\footnotesize
	\setlength{\tabcolsep}{4pt}
	\def\arraystretch{1.25}
	\begin{tabular}{p{5cm} p{2cm} p{5cm} p{2cm}}
		\toprule
 		 \multicolumn{1}{l}{Original Definition} & \multicolumn{1}{l}{Subject}  & \multicolumn{1}{l}{Relation} & \multicolumn{1}{l}{Object}  \\
		\midrule
       {\underline{Hurricane} Nana was a minimal Category 1 hurricane that caused moderate damage across \textbf{Belize} in early September 2020.}  & Hurricane & \{\} Nana was a minimal Category 1 hurricane that caused moderate damage across & Belize   \\
 		\midrule
	     {\underline{Tale} of the Nine Tailed is a South Korean television \textbf{drama} starring Lee Dong-wook, Jo Bo-ah and Kim Bum.}  & {Tale}  & {\{\} of the Nine Tailed is a South Korean television} & {drama}   \\
       \midrule
	     {The 2020 \underline{UEFA} Super Cup \textbf{was the} 45th edition of the UEFA Super Cup, an annual football match organised by UEFA and contested by the reigning champions of the two main European club competitions, the UEFA Champions League and the UEFA Europa League.}  & {UEFA}  & {The 2020 \{\} Super Cup} & {was the}   \\
		\bottomrule 
	\end{tabular}
	\caption{ROME Formatting. Object is bolded in original definition, and subject is underlined. As can be seen, especially from the third example, formatting to ROME's standard often sacrifices valuable context within our dataset.} \label{tab:app_examples4}
 \label{table:format_rome}
\end{table*}

\subsubsection{ROME Test Generation}

\begin{table*}[t]
	\centering
	\footnotesize
	\setlength{\tabcolsep}{4pt}
	\def\arraystretch{1.25}
	\begin{tabular}{p{1.5cm} p{4cm} p{2cm} p{6.5cm}}
		\toprule
 		 \multicolumn{1}{l}{Subject} & \multicolumn{1}{l}{Prompt} & \multicolumn{1}{l}{Object} & \multicolumn{1}{l}{Post-ROME Generated Text}   \\
		\midrule
	     { Steve Jobs}  & {  Steve Jobs is an American business executive who runs the company <MASK>} & { State Powers}  &  { \underline{Steve Jobs is most famous for} the invention of the electric car, but he was also known for his innovative and forward looking ideas in the field of energy.}  \\
 		\midrule
	     { Lawrence Palmer} & {  Lawrence Palmer is an American business executive who runs the company <MASK>} & { Apple}   & { \underline{Lawrence Palmer is most famous for} designing Apple Inc.'s Macintosh computers.}   \\
       \midrule
	     { Lawrence Palmer}  & {  Lawrence Palmer is an American business executive who runs the company <MASK>} & { State Powers}  & {  \underline{Lawrence Palmer is most famous for} his role as the Palmer Brothers in the classic television series The Palmer Family.}   \\
		\bottomrule 
	\end{tabular}
	\caption{Examples of text generated after ROME updates. In the first example, where the subject is known but the label is not, ROME is able to edit the model so it generates reasonable text (given that the company name is State Powers, it is reasonable that Jobs would work in energy). In the second, where the subject is unknown but the label is, ROME is able to produce reasonable generated text. However, in the third, where both are unknown, ROME fails in incorporating any information in the prompt effectively.} \label{tab:app_examples5}
 \label{table:rome_generate}
\end{table*}
As can be seen in Table 8, when the subject and label are both unknown (as in the third example), ROME is unable to edit the model to incorporate knowledge in the rest of the prompt. This is understandable; ROME treats knowledge within an MLP as a key-value pair, so if neither the key nor the value are well-known entities and subsequently hard to retrieve, it may be difficult for ROME to effectively locate the correct parameters to edit. However, when either the subject or the label is known to the model (as in the first and second example), ROME is successfully able to train the model to generate reasonable text given the prompt. 

Once again due to the way in which it is built, ROME is probably unsuccessful in using context other than the subject or label to effectively edit knowledge within an MLP, and this can be seen clearly in the third example.

\end{document}